\documentclass[journal,10pt]{elsarticle}

\usepackage{geometry}
\usepackage{setspace} 
\usepackage{algorithmic}
\usepackage{array}

\usepackage{framed,multirow,booktabs}
\usepackage{graphicx,epsfig}
\usepackage{amssymb,amsmath,bm}
\usepackage{mathrsfs}
\usepackage{textcomp}
\usepackage{amsthm}
\usepackage{lineno,hyperref}
\usepackage{xcolor}
\usepackage{cleveref}
\usepackage[section]{placeins}
\usepackage{pifont}
\newcommand{\cmark}{\ding{51}}%
\newcommand{\xmark}{\ding{55}}%

\journal{Journal of Pattern Recognition}

\begin{document}
	\begin{frontmatter}
		%
		\title{SEMv2: Table Separation Line Detection Based on Instance Segmentation}

		\author{
			Zhenrong Zhang$~^{1,2}$,\quad Pengfei Hu$~^{1}$, \quad Jiefeng Ma$~^{1}$, \quad Jun Du~$^1$, \quad \\
			Jianshu Zhang$^2$, Baocai Yin$^2$, Bing Yin$^2$, Cong Liu$^2$\\
			$^1$ {\em University of Science and Technology of China} \\
			$^2$ {\em iFLYTEK AI Research} \\
		}
		
		\begin{abstract}
			Table structure recognition is an indispensable element for enabling machines to comprehend tables. Its primary purpose is to identify the internal structure of a table. Nevertheless, due to the complexity and diversity of their structure and style, it is highly challenging to parse the tabular data into a structured format that machines can comprehend. In this work, we adhere to the principle of the split-and-merge based methods and propose an accurate table structure recognizer, termed SEMv2 (SEM: \textbf{S}plit, \textbf{E}mbed and \textbf{M}erge). Unlike the previous works in the ``split'' stage, we aim to address the table separation line instance-level discrimination problem and introduce a table separation line detection strategy based on conditional convolution. Specifically, we design the ``split'' in a top-down manner that detects the table separation line instance first and then dynamically predicts the table separation line mask for each instance. The final table separation line shape can be accurately obtained by processing the table separation line mask in a row-wise/column-wise manner. To comprehensively evaluate the SEMv2, we also present a more challenging dataset for table structure recognition, dubbed iFLYTAB, which encompasses multiple style tables in various scenarios such as photos, scanned documents, etc. Extensive experiments on publicly available datasets (e.g. SciTSR, PubTabNet and iFLYTAB) demonstrate the efficacy of our proposed approach. {The code and iFLYTAB dataset are available at \url{https://github.com/ZZR8066/SEMv2}.}
		\end{abstract}
		
		
	\end{frontmatter}
	{\bf Correspondence:}
	Dr. Jun Du, National Engineering Research Center of Speech and Language Information Processing (NERC-SLIP), University of Science and Technology of China, No. 96, JinZhai Road, Hefei, Anhui P. R. China (Email: jundu@ustc.edu.cn).
	
	
	\newpage
	
	\section{Introduction}
	\begin{figure}[t]
		\centerline{\includegraphics[width=1.\linewidth]{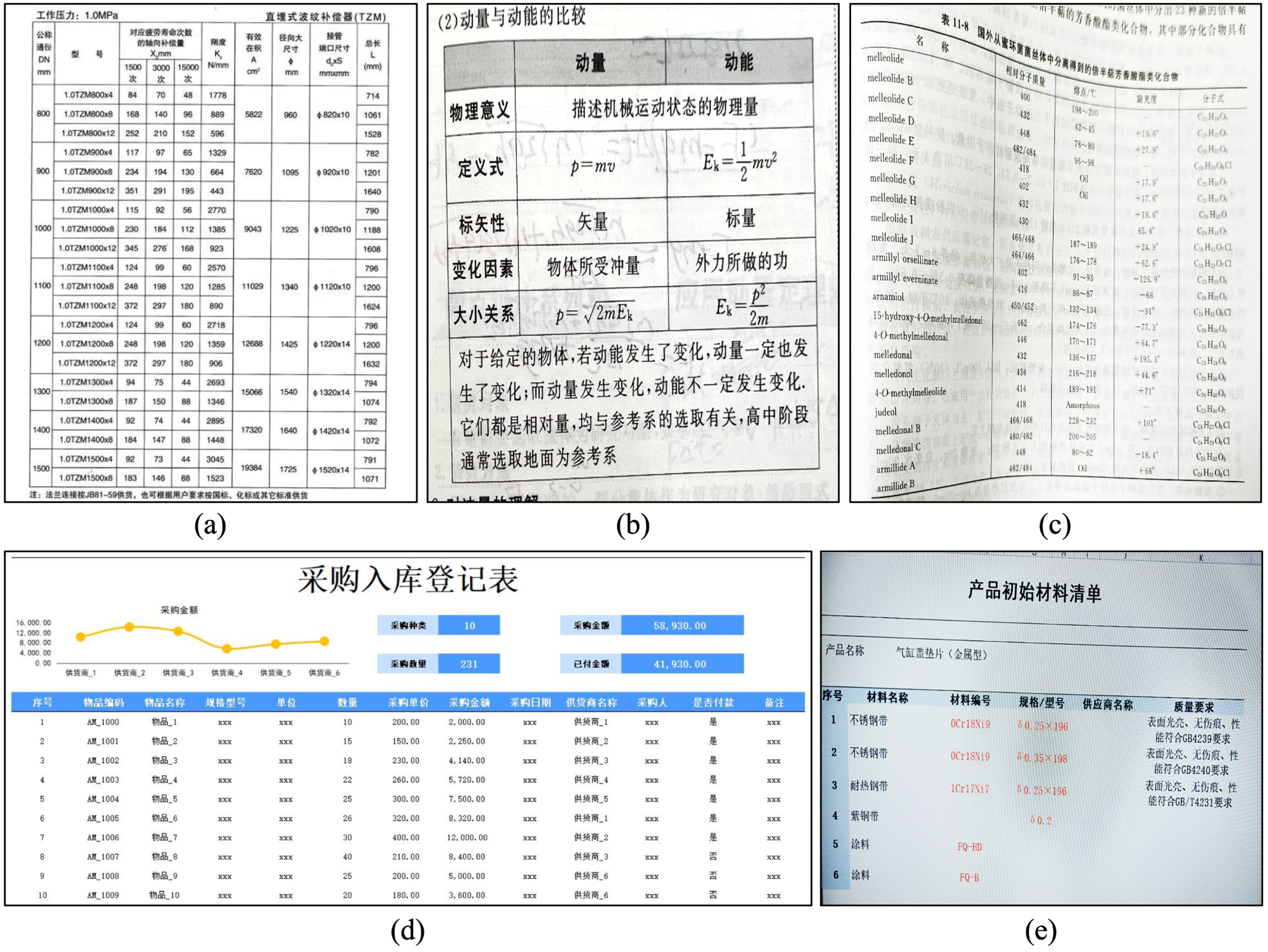}}
		\caption{Some table samples in the iFLYTAB dataset. (a)-(b) are wired tables. (c)-(e) are wireless tables.}
		\label{table_sample}
	\end{figure}
	
	In this era of knowledge and information, document is a significant source of information for numerous cognitive processes such as knowledge database creation, optical character recognition (OCR), document retrieval, etc. As a particular entity, the tabular structure is very commonly encountered in documents. These tabular structures convey important information in a concise form. They are highly prevalent in domains such as finance, administration, research, and even archival documents. Table structure recognition (TSR) aims to recognize the table internal structure to the machine readable data mainly presented in two formats: logical structure and physical structure~\cite{SurveyTable}. More precisely, logical structure only contains every cell’s row and column spanning information, while the physical one additionally contains bounding box coordinates of cells. Therefore, TSR as a precursor to contextual table understanding will be beneficial in a wide range of applications~\cite{DeCNT,DeepDeSRT}.
	
	Limited by the training datasets~\cite{Icdar13,SciTSR,EDD,FinTabNet} used for TSR, most previous works~\cite{SPLERGE,LGPMA,SEMv1,DeepDeSRT} focus on document images that are obtained from digital documents (e.g., PDF files). In such a scenario, the table images are cropped under optimal imaging conditions and are often horizontally (or vertically) aligned with a clean background and distinct table structures. However, in some real-world applications, document images may be captured by mobile cameras. Many camera-captured document images are of poor image quality, and tables contained in them may be distorted (even curved) or contain noises, which makes TSR even more challenging. Although the WTW dataset proposed recently~\cite{WTW} contains table images from natural scenes, it only focuses on wired tables. Parsing wireless tables is a relatively more difficult task due to the lack of visual cues to delimit cells, columns and rows. To comprehensively evaluate the performance of TSR, we present a large-scale dataset in this paper, dubbed iFLYTAB. As shown in Figure~\ref{table_sample}, the table images in the iFLYTAB dataset are collected from various scenarios, and contain both wired and wireless tables.
	
	Considering that a table is composed of a set of table cells and each table cell is composed of one or more basic table grids, the recently proposed split-and-merge based methods~\cite{SPLERGE,SEMv1,RobustTabNet} consider table grids as the fundamental processing units. These methods recognize the table structure as the following pipeline: 1) split table into basic table grid pattern 2) merge grid elements to recover table cells that span multiple rows or columns. When the TSR is performed in this way, once the ``split'' stage predicts erroneous results, it is difficult for the ``merge'' stage to rectify them. Therefore, it is essential to make the model detect table grids more accurately. The previous methods~\cite{SPLERGE,SEMv1,RobustTabNet} complete the first stage in a bottom-up manner. Specifically, they first apply semantic segmentation~\cite{FCN} to predict table row/column separation lines, and then represent the intersection of detected row/column separation lines as table grids. However, segmenting table row/column separation lines in a pixel-wise manner is imprecise due to the limited receptive field. In addition, it necessitates complex mask-to-line algorithms to extract the table separation lines from the predicted segmentation results.
	
	In this work, we follow the split-and-merge based method SEM~\cite{SEMv1}, and introduce an accurate table structure recognizer, termed SEMv2. Distinct from previous segmentation-based methods~\cite{SEMv1,DeepDeSRT,SPLERGE,RobustTabNet} in the ``split'' stage, we aim to distinguish each table separation line and formulate table separation line detection as an instance segmentation task. Specifically, the table separation line mask generation is decoupled into a mask kernel prediction and a mask feature learning, which are responsible for generating convolution kernels and the feature maps to be convolved with respectively. Accurate table row/column separation lines can be easily obtained by processing table row/column separation line masks in a column-wise/row-wise manner. Moreover, compared to the sequence decoder in the ``merge'' stage in~\cite{SEMv1}, we propose a parallel decoder based on conditional convolution to process the merging of basic table grids, which increases the decoding speed. To comprehensively evaluate the SEMv2, we also introduce a new large-scale TSR dataset iFLYTAB, which contains multiple style tables in several scenes like photos, scanned documents, etc.
	
	The main contributions of this paper are as follows:
	
	\begin{itemize}
		
		\item 
		Following the split-and-merge based methods, we propose the SEMv2, which introduces a novel instance segmentation framework for the table separation line detection in the ``split'' stage, making the ``split'' more robust in various scenes.
		
		\item
		We release the iFLYTAB dataset, which is collected from various scenarios and manually annotated carefully, to the community for advancing related research.
		
		\item 
		Based on our proposed method, we achieve state-of-the-art performance on publicly available datasets SciTSR, PubTabNet and iFLYTAB.
		
	\end{itemize}
	
	\section{Related Work}
	\subsection{Existing Datasets}
	{Early datasets for addressing TSR include UW-3~\cite{Uw3}, UNLV~\cite{Unlv}, ICDAR-2013~\cite{Icdar13}, ICDAR-2019~\cite{Icdar19} and TabStructDB~\cite{TabStructDB}. However, the magnitude of these datasets is limited.
	To meet the requirement of data-driven approaches for TSR, large-scale datasets such as Table2Latex~\cite{table2latex}, TableBank~\cite{TableBank} and PubTabNet~\cite{EDD} are proposed, but incomplete annotations still impede their development.
	For instance, TableBank collects 145,463 training tables from the Word and Latex documents. Each table in TableBank solely presents its corresponding HTML tag sequence, devoid of any physical coordinate information.
	Recently, FinTabNet~\cite{FinTabNet}, SciTSR~\cite{SciTSR} and PubTables-1M~\cite{PubTables-1M} add the cell coordinates and row-column information to become relatively comprehensive datasets for TSR.
	Of particular significance is PubTables-1M, which collects nearly a million fully annotated tables sourced from scientific articles. These encompass comprehensive information for table detection, recognition and functional analysis (such as column headers, projected rows and table cells).
	Due to the inconsistency in annotations among these datasets, the efficacy of the model is compromised. \cite{aligndataset} also aligns these benchmark datasets through removing both errors and inconsistency between them, which improves model performance.
	Although dataset scale has been significantly increased, these datasets solely focus on digital documents (e.g., PDF files). 
	Recently, the WTW~\cite{WTW} dataset is introduced, which contains tables in multiple real scenes. {However, it mainly focuses on wired tables, ignoring the more challenging wireless ones.} To comprehensively evaluate the TSR performance, we present a new large-scale dataset iFLYTAB. Different from WTW, tables in iFLYTAB encompass both wired and wireless tables in various scenarios.}
	
	\subsection{Table Structure Recognition}
	Due to the rapid development of deep learning in documents, many deep learning-based TSR approaches~\cite{EDD,GraphTSR,SEMv1,TabStructNet} have been presented. 
	These methods can be roughly divided into three categories: bottom-up methods, image-to-markup based methods and split-and-merge based methods.
	
	One group of bottom-up methods~\cite{TSRNet, GraphTSR,GnnTSR,Res2tim,NCGM} treat words or cell contents as nodes in a graph and use graph neural networks to predict whether each sampled node pair is in the same cell, row, or column. These methods rely on the assumption that the bounding boxes of words or cell contents are available as additional inputs, which are not easy to obtain from table images directly. To eliminate this assumption, another group of methods~\cite{TabStructNet,LGPMA,TOD-Net,CTUNet,FLAG,GTE} proposed to detect the bounding boxes of table cells directly. After cell detection, they designed some rules to cluster cells into rows and columns. However, these methods regard the cells as bounding box, which is difficult to handle the cells in distorted tables. Other methods~\cite{LORE,TGRNet,Cycle-CenterNet} detect cells through detecting the corner points of cells. they can more suitable for distorted cells, but they suffer from tables containing a lot of empty cells and wireless tables.
	
	The image-to-markup based methods~\cite{EDD,table2latex,TableFormer,WSTabNet,VAST} treat table structure recognition as a task similar to image-to-markup generation and directly generate the markup tags that define the structure of the table through an attention-based structure decoder. These methods rely on a large amount of training data and are inefficient as the number of table cells increases.
	
	The split-and-merge based methods~\cite{SPLERGE,SEMv1} first split a table into the basic table grid pattern, and then merge grid elements to recover table cells. Previous methods~\cite{SPLERGE,SEMv1} utilize semantic segmentation~\cite{FCN} for identifying rows, columns within tables in the ``split'' stage. However, segmenting table row/column separation lines in a pixel-wise manner is inaccurate due to the limited receptive field, and heuristic mask-to-line modules designed with strong assumptions in split stage make these methods work only on tables in digital documents. To more accurately split table grids even in distorted tables, RobustTabNet~\cite{RobustTabNet} uses a spatial CNN-based separation line predictor to propagate contextual information across the entire table image in both horizontal and vertical directions. TSRFormer with SepRETR~\cite{TSRFormerv1} formulates the table separation line prediction as a line regression problem and regresses separation line by DETR, but it can't regress too long separation line well. TSRFormer with DQ-DETR~\cite{TSRFormerv2} progressively regresses separation lines, which further enhances localization accuracy for distorted tables. GrabTab~\cite{GrabTab} flexibly fuses multiple components to robustly predicate cell edges. TRACE~\cite{TRACE} first segments the cell corners, explicit lines and implicit lines, then obtains the grid lines from the segmentation result by post-processing, but it needs detailed labels to supervise the train progress which are not available in public datasets. In our work, we formulate the table separation line detection as the instance segmentation task. The table separation line can be accurately obtained by processing the table separation line mask in a row-wise/column-wise manner. 
	
	\subsection{Instance Segmentation}
	Instance segmentation is a challenging task, as it necessitates instance-level and pixel-level predictions simultaneously. The dominant framework for instance segmentation is Mask R-CNN~\cite{MaskRCNN}, which first detects the bounding boxes of objects and then segments the object in the box. Many works~\cite{PANet,HybridRCNN,MaskSRCNN} with top performance are built on Mask R-CNN. Due to the slender shape of table separation lines, this widely utilized box-anchor based instance segmentation methods cannot be employed directly. Another approach to instance segmentation is based on dynamic filter network~\cite{dfn}. For example, SOLOv2~\cite{Solov2} and CondInst~\cite{CondInst} learn instance-dependent convolutional kernels, which are applied to generate instance masks. Inspired by CondInst, we aim to resolve the table row/column instance-level discrimination problem, and propose the conditional table separation line detection strategy.
	
	\section{iFLYTAB}
	\begin{figure}[t]
		\centerline{\includegraphics[width=.45\linewidth]{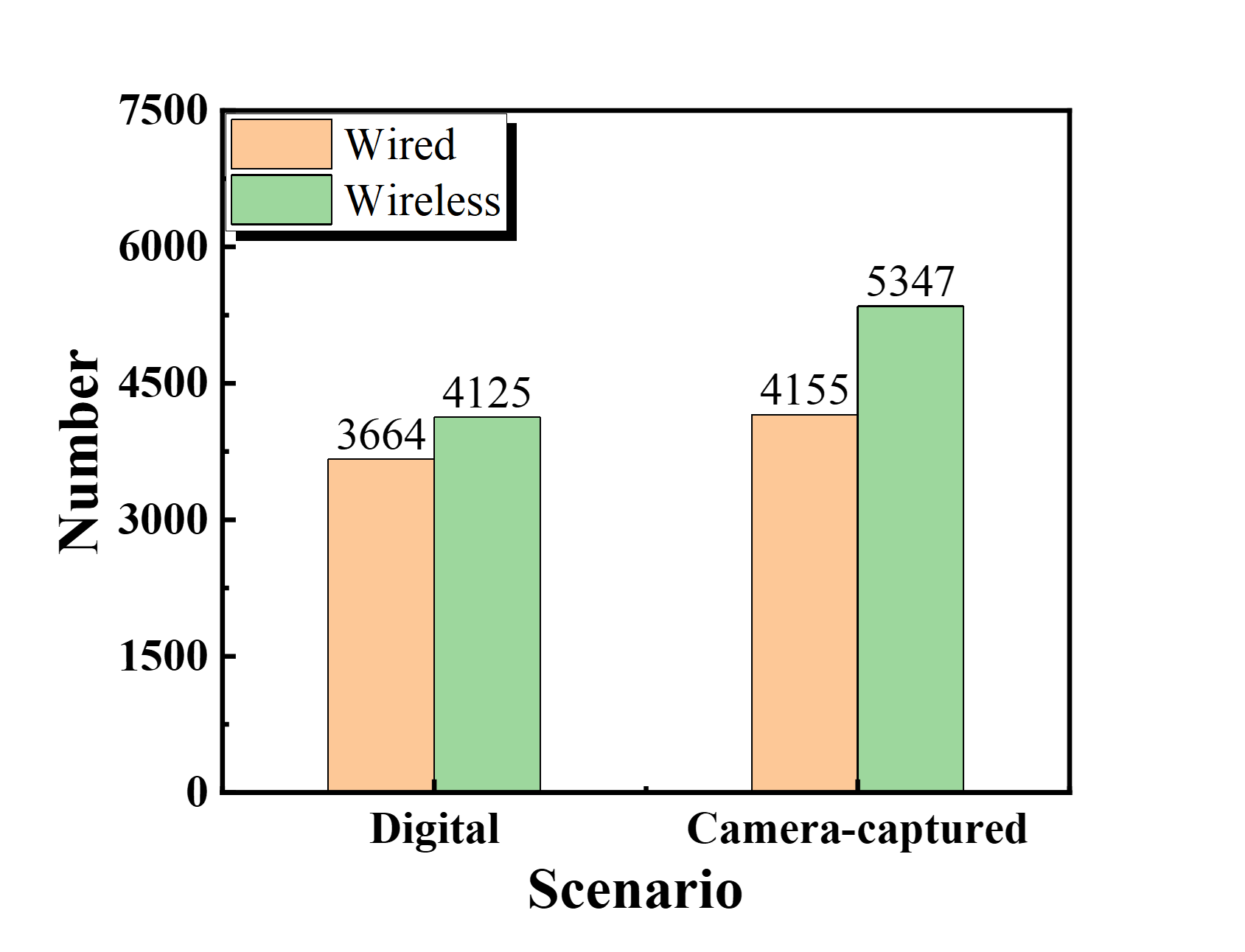}}
		\caption{Statistics of the iFLYTAB datasets.}
		\label{ifltab_statics}
	\end{figure}
	
	\begin{table}[t]
		\centering
		
		\caption{{The comparison between our iFLYTAB
			dataset and the existing datasets for table structure recognition. In the last column, we report
			the total number of samples for all those datasets.}}
		\label{datasets_comparsion}
		\begin{tabular}{lccccc}
			\toprule
			\multirow{2}{*}{Dataset} & \multicolumn{2}{c}{Digital}  & \multicolumn{2}{c}{Camera-captured} & \multirow{2}{*}{Num} \\ \cmidrule(lr){2-3} \cmidrule(lr){4-5}
			& \multicolumn{1}{c}{Wired} & Wireless & \multicolumn{1}{c}{Wired} & Wireless & \\ \midrule
			ICDAR-2013~\cite{Icdar13} & \multicolumn{1}{c}{\cmark} & \cmark & \multicolumn{1}{c}{\xmark}    &  \xmark  & 156 \\ 
			SciTSR~\cite{SciTSR}     & \multicolumn{1}{c}{\cmark} & \cmark & \multicolumn{1}{c}{\xmark}    &  \xmark  & 15,000 \\ 
			TableBank~\cite{TableBank} & \multicolumn{1}{c}{\cmark} & \cmark & \multicolumn{1}{c}{\xmark}    &  \xmark  & 145,000 \\ 
			PubTabNet~\cite{EDD}  & \multicolumn{1}{c}{\cmark} & \cmark & \multicolumn{1}{c}{\xmark}    &  \xmark  & 568,000 \\ 
			FinTabNet~\cite{FinTabNet} & \multicolumn{1}{c}{\cmark} & \cmark & \multicolumn{1}{c}{\xmark}    &  \xmark  & 113,000 \\ 
			PubTables-1M~\cite{PubTables-1M} & \multicolumn{1}{c}{\cmark} & \cmark & \multicolumn{1}{c}{\xmark}    &  \xmark  & 948,000 \\ 
			WTW~\cite{CycleCenterNet}        & \multicolumn{1}{c}{\cmark} &  \xmark   & \multicolumn{1}{c}{\cmark} &   \xmark & 14,581 \\ \midrule
			iFLYTAB    & \multicolumn{1}{c}{\cmark} & \cmark & \multicolumn{1}{c}{\cmark} & \cmark & 17,291 \\ \bottomrule
		\end{tabular}
	\end{table}
	
	iFLYTAB collects table images of various styles from different scenarios. Specifically, as shown in Figure~\ref{ifltab_statics}, we collect both wired and wireless tables from digital documents, and camera-captured photos. As shown in Table~\ref{datasets_comparsion}, compared with existing datasets (e.g. SciTSR, PubTabNet, etc.) that are mainly derived from digital PDF files. the iFLYTAB includes table images captured by cameras, which contain complex image backgrounds and non-rigid image deformation. Although WTW provides table images in the photographic scenario, it ignores the more challenging wireless tables. 
	
	In terms of data labeling, we provide comprehensive annotation for each table image including physical coordinates and row/column information. 
	Subsequently, we will present a detailed exposition on the annotation of physical coordinates and row/column information.
	
	\begin{figure*}[t]
		\centerline{\includegraphics[width=1\linewidth]{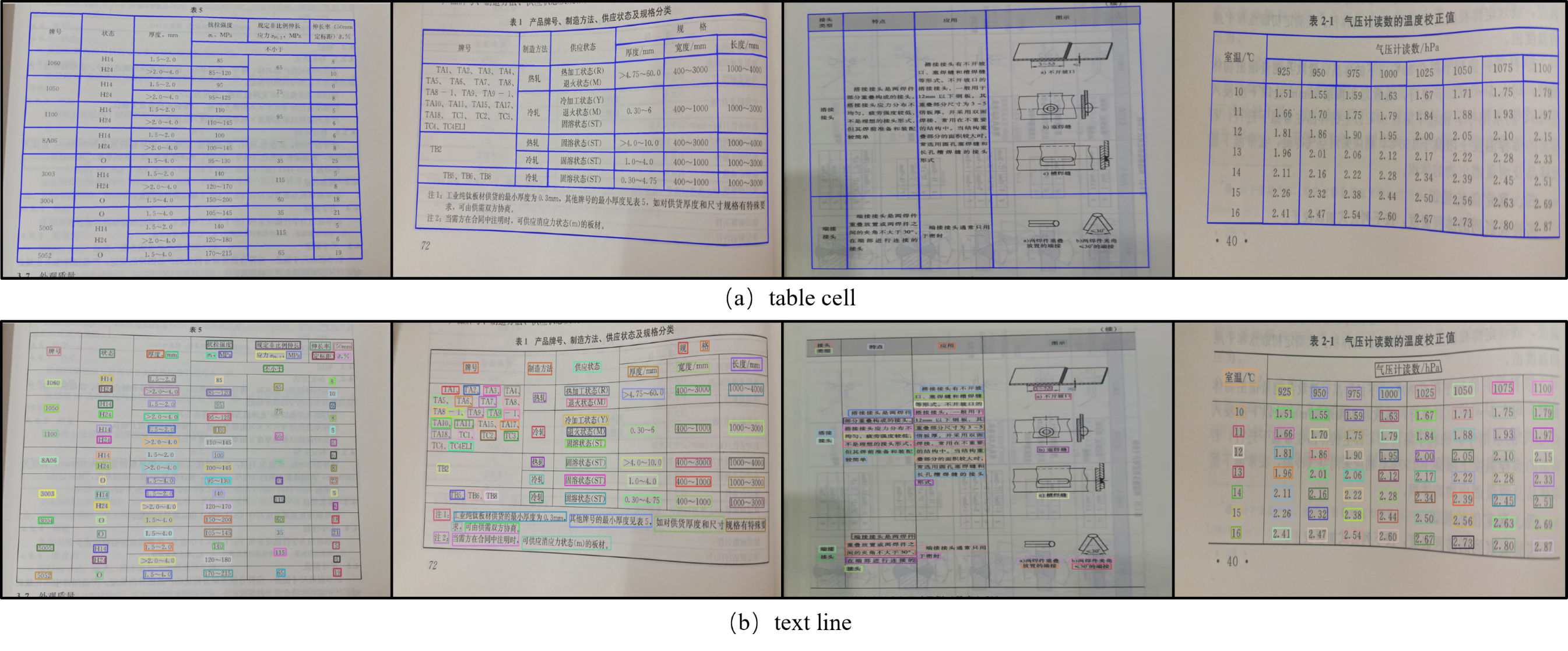}}
		\caption{The visualization of annotated physical coordinates. (a) refers to table cell polygons. (b) refers to text line polygons. Best view in zoom in.}
		\label{physical-coordinat}
	\end{figure*}
	
	\begin{figure*}[t]
		\centerline{\includegraphics[width=1\linewidth]{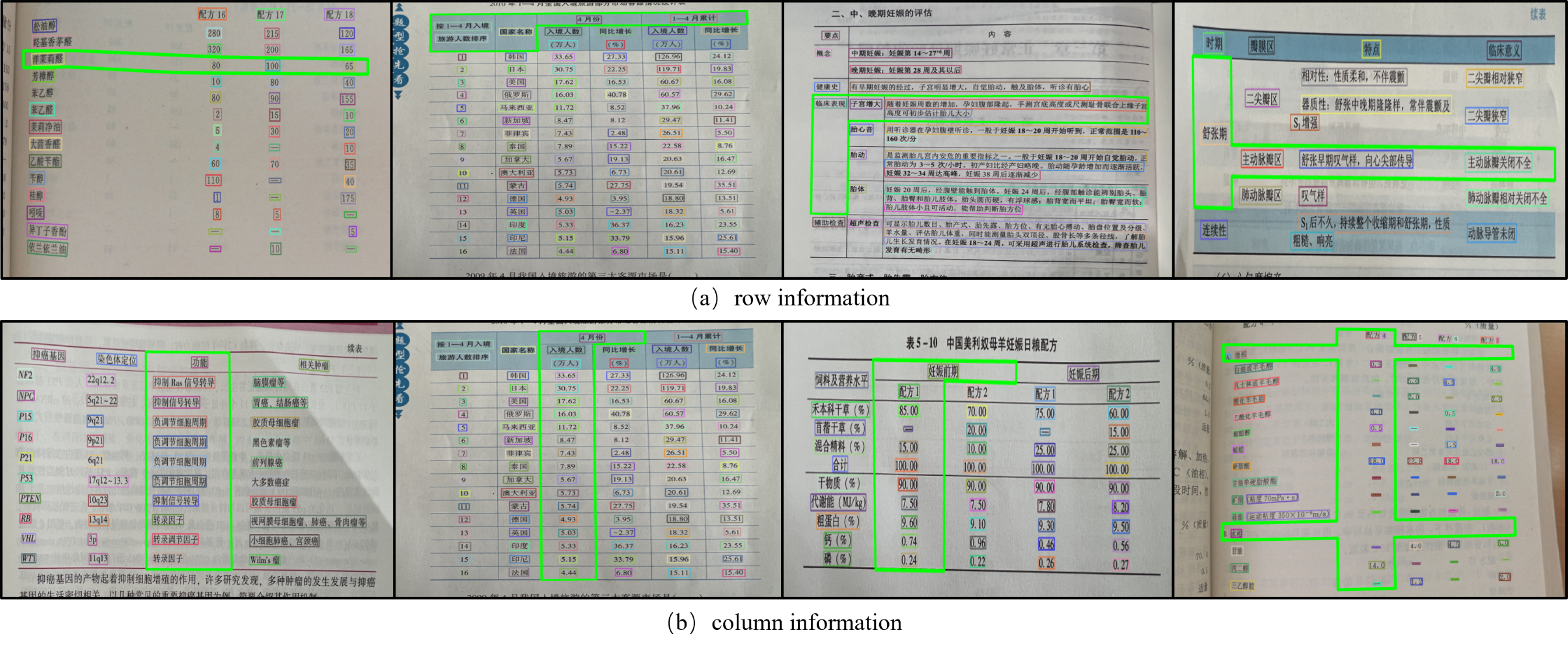}}
		\caption{The visualization of annotated row/column information. Then green polygons are the annotated row/column information. (a) refers to row information. (b) refers to column information. Best view in zoom in.}
		\label{row-col-information}
	\end{figure*}
	
	\textbf{Physical Coordinates} As illustrated in Figure~\ref{physical-coordinat}, the physical coordinates we have annotated comprise of both table cell and text line polygons. 
	Each polygon is labeled as $\{x^{\text{lt}},y^{\text{lt}},x^{\text{rt}},y^{\text{rt}},x^{\text{rb}},y^{\text{rb}},x^{\text{lb}},y^{\text{lb}}\}$, representing the coordinates of the four vertices.
	
	\textbf{Row/column Information} The row/column information is employed to ascertain which text lines are attributed to the same row/column in a table.  Therefore, we additionally provide a series of polygons that envelop text lines belonging to the same row or column. As depicted in Figure~\ref{row-col-information}, the text lines enclosed in the green polygon indicate that they locat at the same row/column in a table.
	
	We have manually annotated 735,781 polygons for table cells, 1,207,709 polygons for text lines, 207,972 polygons for row information, and 112,820 polygons for column information.
	We randomly select approximately 70\% of the table images as the training set, and the rest data samples are used for testing. Finally, our iFLYTAB dataset has 12,104 training samples and 5,187 testing ones.
	
	\begin{figure*}[t]
		\centerline{\includegraphics[width=1.\linewidth]{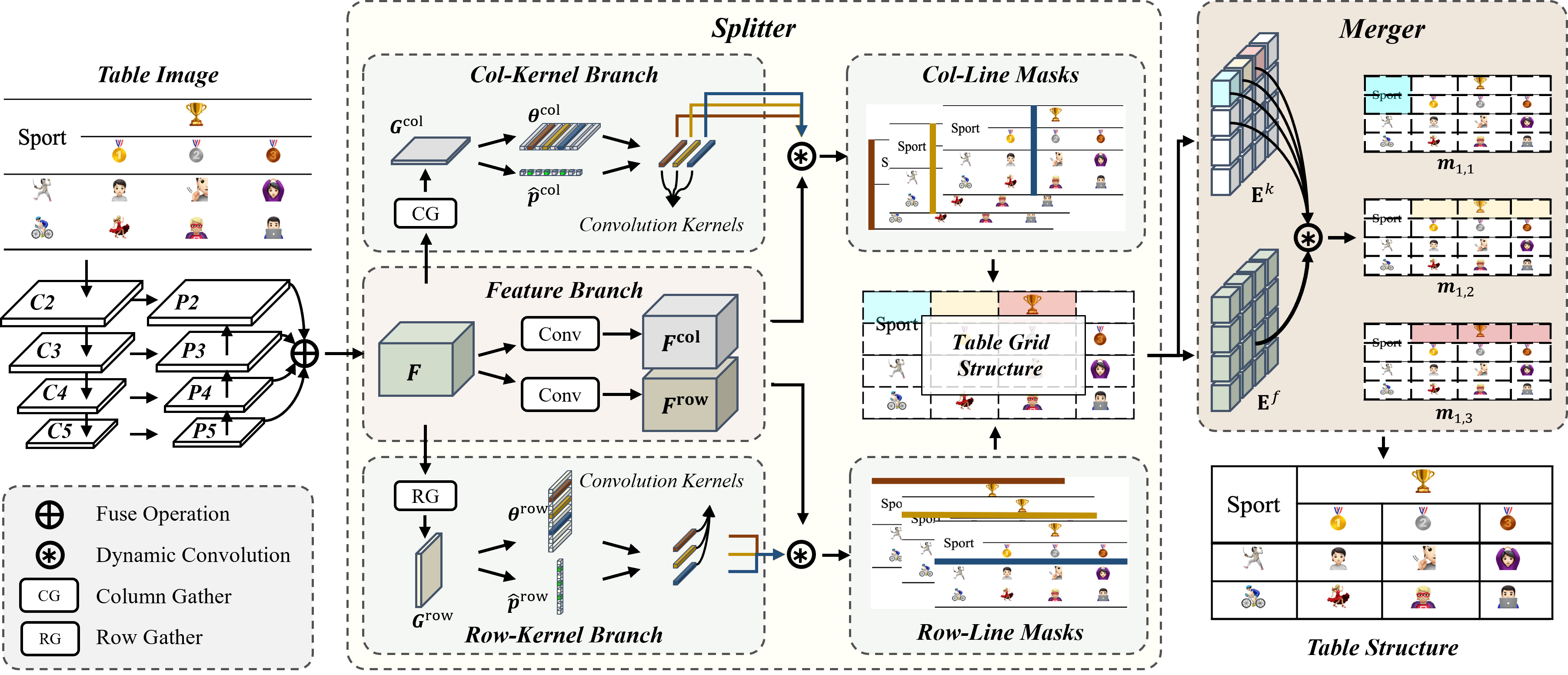}}
		\caption{The overall architecture of SEMv2. $\boldsymbol{F}$ is the feature map generated by fusing the FPN feature maps ($P$2 to $P$5). The \textit{Splitter} module consists of \textit{Kernel Branch} and \textit{Feature Branch}, and predicts table separation lines between different columns or rows, which can be further processed to obtain the \textit{Table Grid Structure}. The \textit{Merger} module predicts the table cell to which each table grid belongs. We omit the \textit{Embedder} module for simplicity.}
		\label{system}
	\end{figure*}
	
	\section{Method}
	The schematic of our approach is depicted in Figure~\ref{system}. SEMv2 adheres to the split-and-merge based methodology of its predecessor, SEM, and is primarily comprised of three components: the splitter, the embedder, and the merger. The splitter takes the table image as input and predicts the fine grid structure of the table. The embedder extracts the grid-level feature representation of each basic table grid. The merger predicts which grids should be merged to recover the whole table structure. In the following sections, we will elucidate each component.
	
	\subsection{Splitter}
	Given an input table image $\boldsymbol{I}\in \mathbb{R}^{H\times W\times 3}$, as illustrated in Figure~\ref{system}, the objective of the splitter is to predict the table grid structure with a set of grid bounding boxes $\boldsymbol{B} \in \mathbb{R}^{M\times N\times 4}$, where $M$, $N$ are the number of rows and columns occupied by the table grid structure respectively. Previous split-and-merge based methods apply the semantic segmentation to predict all table row/column separation lines in one mask and subsequently represent the intersection of detected row/column separation lines as grid bounding boxes $\boldsymbol{B}$. In contrast to prior methods, we formulate table separation line detection as an instance segmentation task and endeavor to predict an individual mask for each table row/column separation line.
	
	The overall architecture of our splitter is depicted in Figure~\ref{system}. The ResNet-34~\cite{ResNet} with FPN~\cite{FPN} is utilized to generate a feature pyramid with four feature maps $\{\boldsymbol{P}_2, \boldsymbol{P}_3, \boldsymbol{P}_4, \boldsymbol{P}_5 \}$, whose scales are 1/4, 1/8, 1/16, 1/32 respectively. To amalgamate the information from all levels of the FPN pyramid into a single output $\boldsymbol{F}$, we also propose a straightforward fuse operation as follows:
	\begin{align}
		\boldsymbol{F}=\boldsymbol{P}_2+\text{Up}_{1\times}\left( \boldsymbol{P}_3 \right) +\text{Up}_{2\times}\left( \boldsymbol{P}_4 \right) +\text{Up}_{3\times}\left( \boldsymbol{P}_5 \right) 
	\end{align} 
	where $\text{Up}_{{n}\times}$ denotes the $n$ times bilinear upsample operation. $\boldsymbol{F}\in \mathbb{R}^{\frac{H}{4}\times \frac{W}{4}\times C}$, where $C$ denotes the number of feature channels.
	
\begin{figure*}[t]
		\centerline{\includegraphics[width=1.\linewidth]{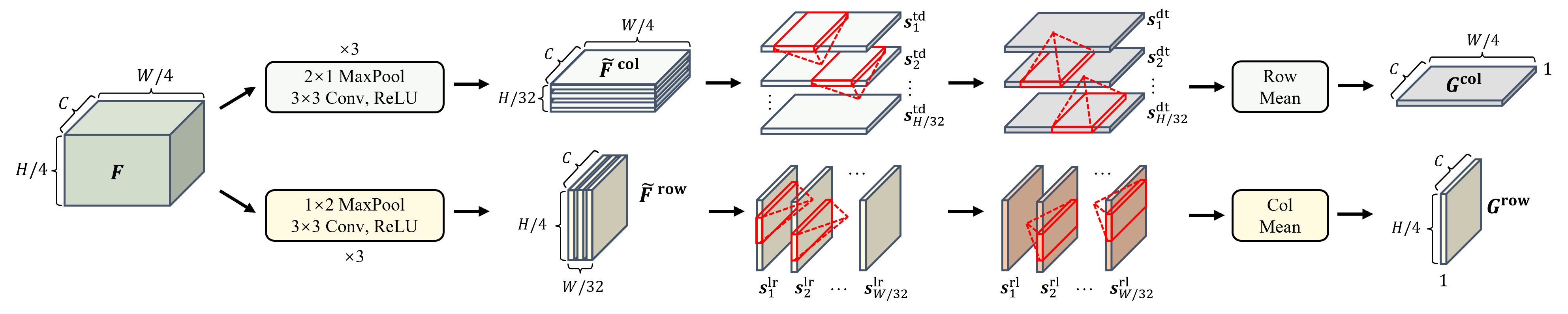}}
		\caption{Illustration of the Gather architecture. The upper part is the Column Gather. The lower part is the Row Gather.}
		\label{gather}
	\end{figure*}

	Inspired by CondInst~\cite{CondInst}, we decouple the table separation line mask generation into a feature branch and a kernel branch. {The feature branch contains two $1 \times 1$ convolution layers for generating $\boldsymbol{F}^{\text{col}}$/$\boldsymbol{F}^{\text{row}}\in \mathbb{R}^{\frac{H}{4}\times \frac{W}{4}\times C}$}, which will be convoluted with convolution kernels from kernel branches to predict separation line masks. Since the table separation lines are usually slender and traverse the entire table image, it is necessary to design a kernel branch that has a broader receptive field. To address this issue, we propose the Gather module to capture the horizontal/vertical visual clues as shown in Figure~\ref{gather}.
	
	Taking the Column Gather as an example, we first conduct three repeated down-sampling operations on $\boldsymbol{F}$, and each operation is composed of a sequence of a $2\times 1$ max-pooling layer, a $3\times 3$ convolutional layer and a ReLU activation function. The down-sampled feature map $\boldsymbol{\tilde{F}}^{\text{col}} \in \mathbb{R}^{\frac{H}{32}\times \frac{W}{4}\times C}$ will be taken as the input of two following spatial CNN modules~\cite{SCNN}. The first spatial CNN module divides the feature map into ${H}/{32}$ slices, which are denoted as $\boldsymbol{S}^{\text{td}}=\left\{ \left. \boldsymbol{s}_{i}^{\text{td}}\in \mathbb{R}^{1\times \frac{W}{4}\times C} \right|i=1,2,...,\frac{H}{32} \right\}$. Specifically, the topmost slice $\boldsymbol{s}^{\text{td}}_1$ is convolved by a $1\times 5$ convolution layer, and its output feature map is merged with the next slice $\boldsymbol{s}^{\text{td}}_2$ by element-wise addition. This	procedure is done iteratively so that the information can be propagated from the topmost to the bottommost effectively. The second spatial CNN module transmits information in a reversed direction. In this way, each pixel in the output feature map can leverage the structural information from both sides to enhance its feature representation ability. $\boldsymbol{G}^{\text{col}}\in \mathbb{R}^{1\times \frac{W}{4}\times C}$ is obtained by taking the row mean of the enhanced feature map. We add a linear transformation following the $\boldsymbol{G}^{\text{col}}$ to predict $C$-dimensional output $\pmb{\theta}^{\text{col}}\in \mathbb{R}^{1\times \frac{W}{4}\times C}$. $\pmb{\theta}^{\text{col}}$ will be used as the weights of a $1\times 1$ convolution layer to predict table column separation line masks. We also detect table column separation line instance by predicting $\boldsymbol{\hat{p}}^{\text{col}} \in \mathbb{R}^{1\times \frac{W}{4}\times 1}$ through a linear transformation. The loss function on $\boldsymbol{\hat{p}}^{\text{col}}$ is formulated as follows:
	\begin{equation}\label{s}
		\mathscr{L}_{\text{inst}}^{\text{col}}=\sum_{i=1}^{W/4}{\frac{L_{\text{bce}}\left( \hat{p}_{i}^{\text{col}},\tilde{p}_{i}^{\text{col}} \right)}{W/4}}
	\end{equation}
	
\begin{figure}[t]
		\centerline{\includegraphics[width=.5\linewidth]{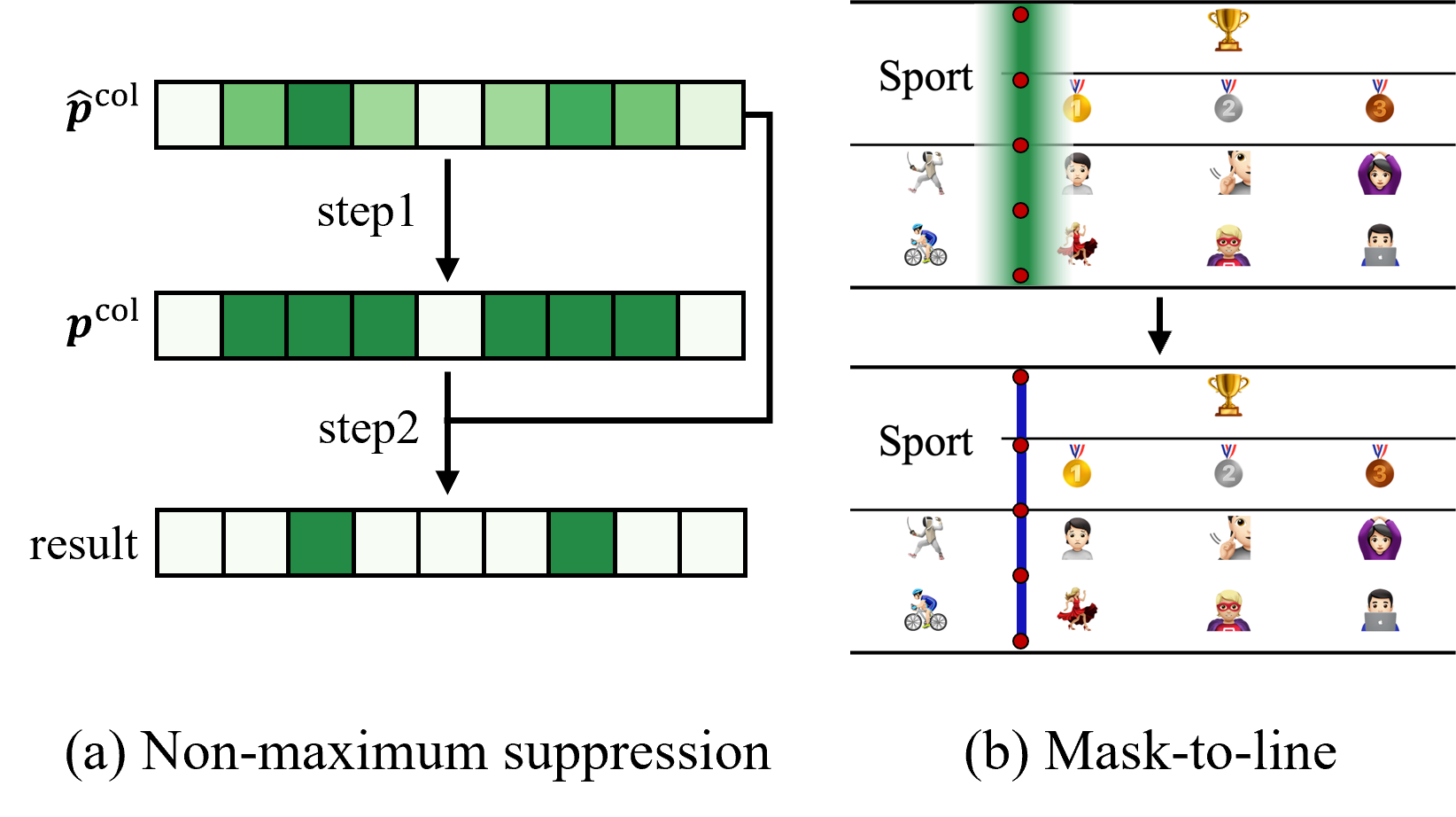}}
		\caption{The illustration of Post-processing.}
		\label{post-processing}
	\end{figure}

	where $L_{\text{bce}}$ is the binary cross-entropy loss, $\tilde{\boldsymbol{p}}^{\text{col}}$ denotes the ground-truth distribution of starting points of table column separation lines on the x-axis. $\tilde{p}^{\text{col}}_i$ is 1 if the start point of a table column separation line is located in the $i$-th column, otherwise 0. To eliminate the duplicated predictions of the starting point of a table column separation line in $\tilde{\boldsymbol{p}}^{\text{col}}$, as shown in Figure~\ref{post-processing}(a), we perform non-maximum suppression as follows: 
	1) binarize the $\hat{\boldsymbol{p}}^{\text{col}}$ into $\boldsymbol{p}^{\text{col}}$, 
	2) for the continuous pixels whose value equals $1$ in $\boldsymbol{p}^{\text{col}}$, the pixel with maximum score in $\boldsymbol{\hat{p}}^{\text{col}}$ will be selected to represent a table column separation line instance.
	
	According to the detected table separation line instance, we select convolution kernels from $\pmb{\theta}^{\text{col}}$ to conduct dynamic convolution with $\boldsymbol{F}^{\text{col}}$ to predict table column separation line masks $\boldsymbol{\hat{F}}^{\text{col}}\in \mathbb{R}^{\frac{H}{4}\times \frac{W}{4}\times N^{\text{col}}}$, where $N^{\text{col}}$ represents the number of detected table column separation line instances. The loss function on $\boldsymbol{\hat{F}}^{\text{col}}$ is defined as follow:
	\begin{equation}
		\mathscr{L}_{s}^{\text{col}}=\frac{1}{N^{\text{col}}}\sum_{k=1}^{N^{\text{col}}}{\sum_{i=1}^H{\sum_{j=1}^W{\frac{L_{\text{f}}\left( \hat{F}_{i,j,k}^{\text{col}},\tilde{{F}}_{i,j,k}^{\text{col}} \right)}{\sum_{i=1}^H{\sum_{j=1}^W{\tilde{{F}}_{i,j,k}^{\text{col}}}}}}}}
	\end{equation}
	in which
	\begin{align}
		L_{\text{f}}\left( x,y \right) =\left\{ \begin{array}{c}
			\alpha \left( 1-\sigma \left( x \right) \right) ^{\gamma}\log \left( \frac{1}{\sigma \left( x \right)} \right) ,if\ y=1\\
			\left( 1-\alpha \right) \sigma \left( x \right) ^{\gamma}\log \left( \frac{1}{1-\sigma \left( x \right)} \right) ,if\ y=0\\
		\end{array} \right. 
		\label{focalloss}
	\end{align}
	where $\boldsymbol{\tilde{F}}^{\text{col}}$ denotes the ground-truth of table column separation line masks. ${\tilde{{F}}}_{i,j,k}^{\text{col}}$ is $1$ if the pixel in $i$-th row, $j$-th column and $k$-th channel belongs to $k$-th table column line, otherwise 0. The function $L_{\text{f}}$ is actually the sigmoid focal loss~\cite{focalloss} and the $\sigma$ is the sigmoid function.
	
	Considering the table column separation lines are typically spread through vertical direction, we process $\boldsymbol{\hat{F}}^{\text{col}}$ in a row-wise manner to obtain the final table column separation line. Specifically, as shown in Figure~\ref{post-processing}(b), given a table column separation line mask, we first find the maximum score of each row, which is presented as a set of red dots as shown in Figure~\ref{post-processing}(b). The table column separation line can be obtained by connecting these red dots together. The grid bounding boxes $\boldsymbol{B}$ can be derived from the intersection of table row/column separation lines.
	
	\subsection{Embedder}
	
\begin{figure}[t]
		\centerline{\includegraphics[width=.75\linewidth]{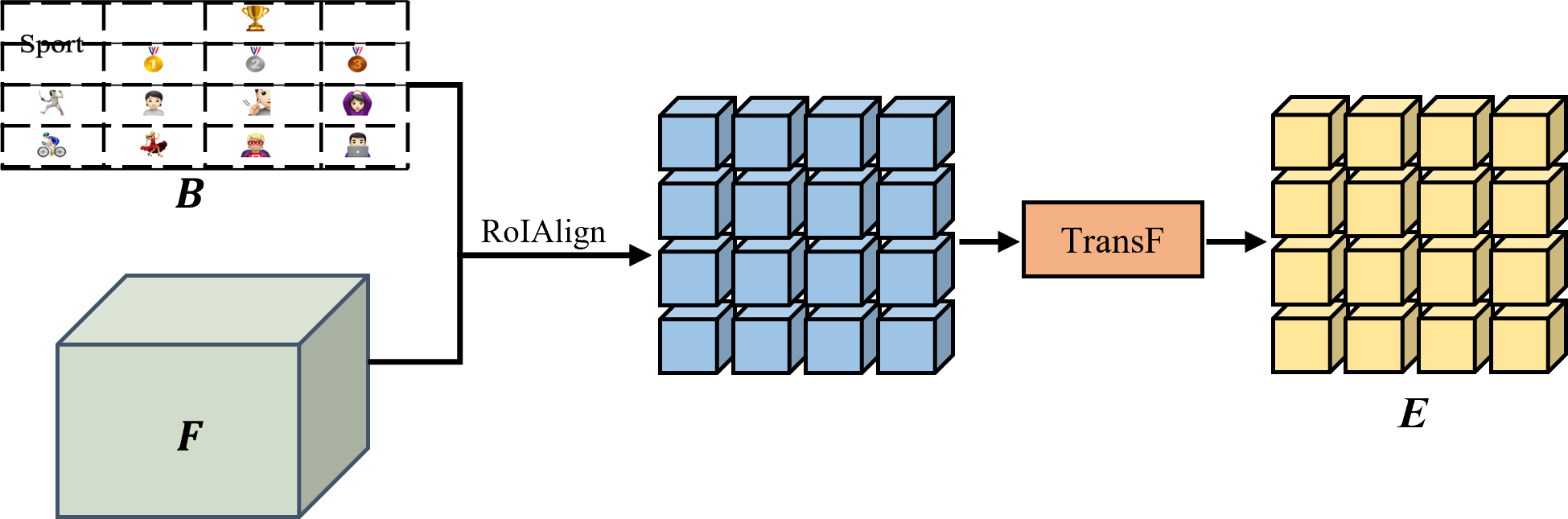}}
		\caption{{The illustration of the embedder. The embedder extracts the gird-level features from $\boldsymbol{F}$ according to table grid boxes $\boldsymbol{B}$. Then the transformer module is applied to the grid-level feature to obtain the final grid-level feature $\boldsymbol{E}$. ``TransF'' denotes the transformer module.}}
		\label{embedder}
	\end{figure}

	The embedder aims to extract the grid-level feature representations $\boldsymbol{E}\in \mathbb{R}^{ M\times N \times D}$, where $D$ is the number of feature channels. As shown in Figure~\ref{embedder}, we take the image-level feature map $\boldsymbol{F}$ and the well-divided table grids $\boldsymbol{B}$ obtained from the splitter as input, and apply the RoIAlign~\cite{MaskRCNN} to extract a fixed size $R\times R$ feature map $\boldsymbol{\hat{e}}_{i,j} \in \mathbb{R}^{R\times R \times C}$ for each grid.
	\begin{align}
		\boldsymbol{\hat{e}}_{i,j}=\text{RoIAlign}_{R\times R}\left( \boldsymbol{F,b}_{i,j} \right)
	\end{align}
	Then two linear transformations with a ReLU function are conducted on $\boldsymbol{\hat{e}}_{i,j}$ to obtain the $D$-dimension output:
	\begin{align}
		\boldsymbol{e}_{i,j}=\max \left( 0,\boldsymbol{\hat{e}}_{i,j}\boldsymbol{W}_1+\boldsymbol{b}_1 \right) \boldsymbol{W}_2+\boldsymbol{b}_2 
	\end{align}
	where $\boldsymbol{W}_1$ and $\boldsymbol{W}_2$ are learned projection matrices, $\boldsymbol{b}_1$ and $\boldsymbol{b}_2$ are learned biases. So far, the features of each basic table grid are still independent of each other. Therefore, we introduce the transformer~\cite{Attention} to capture long-range dependencies on table grid elements and utilize its output as the final grid-level features $\boldsymbol{E}$. 
	
	\subsection{Merger}
	
\begin{figure}[t]
		\centerline{\includegraphics[width=.75\linewidth]{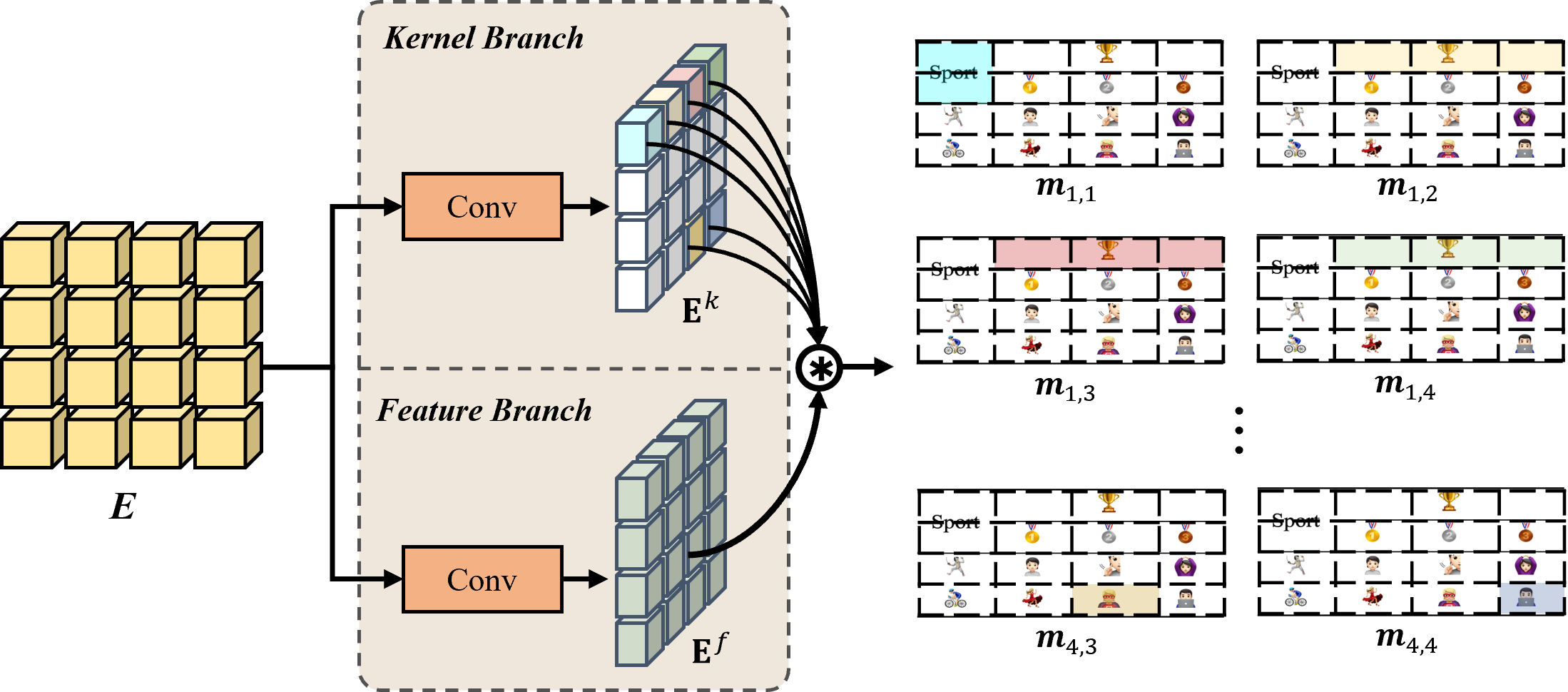}}
		\caption{{The illustration of the merger. The merger first generates the feature maps $\boldsymbol{E}^f$ and kernel parameters $\boldsymbol{E}^k$ through a feature branch and a kernel branch, respectively. Then we convolute the feature map $\boldsymbol{E}^f$ with kernel parameter $\boldsymbol{e}^k_{i,j}$, which is utilized as the weights of a $1\times 1$ convolution layer, to obtain the merged map $\boldsymbol{m}_{i,j}$ of the $i$-th row, $j$-th column grid.}}
		\label{merger}
	\end{figure}

	The merger takes the grid-level features $\boldsymbol{E}$ as input and yields a set of merged maps, which can be formulated as: $\boldsymbol{M}=\left\{ \boldsymbol{m}_{1,1},\cdots
	,\boldsymbol{m}_{M,N} \right\} ,\boldsymbol{m}_{i,j}\in \left\{ 0,1 \right\} ^{M\times N},i\in \left\{1,\cdots ,M \right\}, j\in \left\{1,\cdots ,N \right\}$. Following the approach of the splitter, as illustrated in Figure~\ref{merger}, we use a feature branch and a kernel branch to predict $\boldsymbol{M}$ jointly. Each branch contains only one $1\times 1$ convolution layer to generate feature maps $\boldsymbol{E}^f\in \mathbb{R}^{M\times N\times D}$ and kernel parameters $\boldsymbol{E}^k\in \mathbb{R}^{M\times N\times D}$. We first convolute the feature map $\boldsymbol{E}^f$ with kernel parameter $\boldsymbol{e}^k_{i,j}$, which is utilized as the weights of a $1 \times 1$ convolution layer, to predict the merged map $\boldsymbol{\hat{m}}_{i,j}\in \left[ 0,1 \right] ^{M\times N}$. The loss function is formulated as follows:
	\begin{equation}
		\mathscr{L}_m=\frac{1}{M\times N}\sum_{i=1}^M{\sum_{j=1}^N{\frac{L_{\text{f}}\left( \boldsymbol{\hat{m}}_{i,j},\boldsymbol{\tilde{m}}_{i,j} \right)}{\lVert \boldsymbol{\tilde{m}}_{i,j} \rVert _1}}}
	\end{equation}
	where $\lVert \cdot \rVert _1$ is the L1 norm, the function $L_{\text{f}}$ has been defined in Eq.~(\ref{focalloss}), $\tilde{\boldsymbol{m}}_{i,j}\in \left\{ 0,1 \right\} ^{M\times N}$ denotes the merged map ground-truth of the $i$-{th} row, $j$-{th} column table grid. If the value of $\tilde{\boldsymbol{m}}_{i,j}$ is equal to $1$, then it indicates that the corresponding grid is associated with the identical table cell; otherwise, 0. The final merged map $\boldsymbol{m}_{i,j}$ can be obtained by binarizing $\boldsymbol{\hat{m}}_{i,j}$.
	
	\section{Experiment}
	{In this section, we perform comprehensive experiments on the SciTSR~\cite{SciTSR}, PubTabNet~\cite{EDD}, cTDaR~\cite{Icdar19}, WTW~\cite{WTW} and the proposed iFLYTAB dataset to verify the effectiveness of SEMv2.} 
	We first introduce the relevant datasets and metrics, and then demonstrate the label generation and implementation details of our method.
	Additionally, we visualize the predicted results of our model and conduct ablation studies to analyze the effectiveness of our proposed splitter, gather and merger. 
	Our model is compared with state-of-the-art methods on public benchmark datasets.
	
	\subsection{Datasets}
	\textbf{SciTSR} comprises of 12,000 training samples and 3,000 testing samples of axis-aligned tables extracted from the scientific literature.
	Furthermore, to reflect the model’s ability of recognizing complex tables, it also selects all the 716 complex tables from the test set as a more challenging test subset, called SciTSR-COMP.
	It is worth noting that the test set of SciTSR encompasses the presence of annotation errors.
	To ensure a more accurate evaluation of the model's performance, we follow the RobusTabNet~\cite{RobustTabNet} and rectify the annotation errors present within the test set.
	Thus, it is pertinent to acknowledge that a comparison with other methods on the SciTSR dataset may entail a degree of unfairness, while the comparison on PubTabNet will emerge as a more equitable benchmark.
	
	\textbf{PubTabNet} is a large-scale table recognition dataset, which contains 500,777 training samples and 9,115 validating samples. 
	PubTabNet annotates each table image with information about both the structure of table and the text content with position of each non-empty table cell.
	All tables are also axis-aligned and collected from scientific articles.
	
	{\textbf{cTDaR TrackB1-Historical~\cite{Icdar19}} contains 600 training samples and 150 testing samples. It is worth noting that the table images utilized in this dataset are historical documents with handwritten. This dataset provides the physical coordinates and structure information for each table cell.}
	
	{\textbf{WTW} contains 10,970 training images and 3,611 testing images collected from wild complex scenes. This dataset focuses on wired tabular objects only and provides the annotated information of tabular cell coordinates, and row/column information.}
	
	\textbf{iFLYTAB} is the proposed dataset in this paper, which contains 12,104 training samples and 5,187 testing samples. 
	We provide comprehensive annotation for each table image including physical coordinates and structure information.
	However, it is worth noting that iFLYTAB does not provide annotations for the textual content within table images.
	In addition to the axis-aligned digital documents, the collected table images also include images taken by cameras, which are more challenging due to the complicated background and non-rigid image deformation.
	
	\subsection{Metric}
	{We use F1-Measure~\cite{F1-Measure}, Tree-Edit-Distance-based Similarity (TEDS)~\cite{EDD}, WAvg.F1~\cite{Icdar19} and GriTS~\cite{grits} metric, which are commonly adopted in table structure recognition literature and competitions, to evaluate the performance of our model for recognition of the table structure.}
	
	In order to use the F1-Measure, the adjacency relationships among the table cells need to be detected. 
	F1-Measure measures the percentage of correctly detected pairs of adjacent cells, where both cells are segmented correctly and identified as neighbors. 
	{When evaluating on the WTW dataset, we use the cell
		adjacent relationship metric~\cite{wtw-eval}. This metric is a variant of F1-Measure that maps a groundtruth cell with a predicted cell according to the Intersection over Union (IoU). Here we use IoU=0.6.}
	
	TEDS measures the similarity of the tree structure of tables. 
	While using the TEDS metric, we need to present tables as a tree structure in the HTML format. 
	Finally, TEDS between two trees is computed as:
	\begin{equation}
		\text{TEDS}(T_a, T_b) = 1 - \frac{\text{EditDist}(T_a, T_b)}{\max (\lvert T_a \rvert, \lvert T_b \rvert)}
	\end{equation}
	where $T_a$ and $T_b$ are the tree structure of tables in the HTML formats. 
	EditDist represents the tree-edit distance~\cite{TED}, and $\lvert T \rvert$ is the number of nodes in $T$.
	
	Since taking OCR errors into account may lead to an unfair comparison due to the different OCR models used by various TSR methods, we also employ a modified version of TEDS, called TEDS-Struct. The TEDS-Struct assesses the accuracy of table structure recognition, while disregarding the specific outcomes generated by OCR.
	
	{While using the WAvg.F1, the precision, recall and F1 value of cell detection and cell pair relation prediction need to be calculated at IoU thresholds of [0.6, 0.7, 0.8, 0.9]. The weighted average F1 (WAvg.F1) value of all IoU thresholds is defined as:
		\begin{equation}
			\text{WAvg.F}1=\frac{\sum_{i=1}^4{\text{IoU}_i\cdot \text{F1@IoU}_i}}{\sum_{i=1}^4{\text{IoU}_i}}
		\end{equation}
	}
	
	{The recently proposed GriTS metric compares predicted tables and ground truth directly in matrix form and can be interpreted as an F-score over the correctness of predicted cells. Exact match accuracy considers the percentage of tables for which all cells, including blank cells, are matched exactly.}
	
	In our experiments, we align the official provided text contents to the predicted table cells according to the IOU metric. 
	Ultimately, the output for each table image encompasses the physical coordinates of predicted cell bounding boxes, accompanied by spanning information its corresponding content.
	It is worth noting that the iFLYTAB dataset does not provide text content annotation. 
	Consequently, during the evaluation of iFLYTAB, we assign a distinctive marker to each text line, signifying its individual content. 
	The evaluation code has been made available to the public and can be accessed at the following link: \footnote{\label{dataprocess}\url{https://github.com/ZZR8066/SEMv2}}.
	
	\subsection{Label Generation}
	\textbf{Label of Splitter}
	Distinct from previous methods, we formulate table separation line detection as an instance segmentation task and endeavor to predict an individual mask for each table row/column separation line. 
	Two labels necessitate generation to guide the training of the splitter, namely the table row/column separation line masks $\boldsymbol{\tilde{F}}^{\text{row}}$/$\boldsymbol{\tilde{F}}^{\text{col}}$ and table row/column line instance $\tilde{\boldsymbol{p}}^{\text{row}}$/$\tilde{\boldsymbol{p}}^{\text{col}}$. 
	
\begin{figure}[t]
		\centerline{\includegraphics[width=1\linewidth]{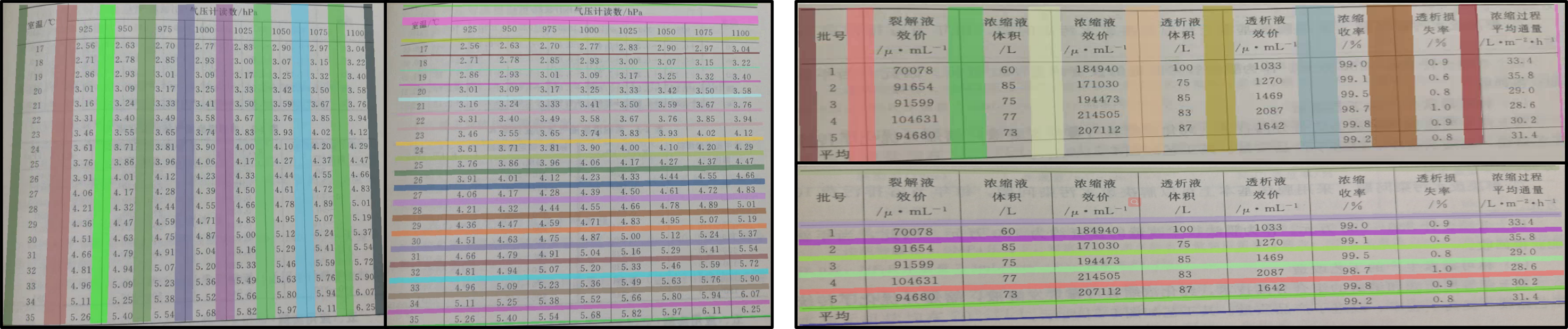}}
		\caption{The visualization of generated table row/column separation line mask labels $\boldsymbol{\tilde{F}}^{\text{row}}$/$\boldsymbol{\tilde{F}}^{\text{col}}$.}
		\label{row-col-line-masks}
	\end{figure}

	Following the SEM~\cite{SEMv1}, the table separation line mask $\boldsymbol{\tilde{F}}^{\text{row}}$/$\boldsymbol{\tilde{F}}^{\text{col}}$ are designed to maximize the size of the separator regions without intersecting any non-spanning cell content, as shown in Figure~\ref{row-col-line-masks}. 
	
\begin{figure}[t]
		\centerline{\includegraphics[width=1\linewidth]{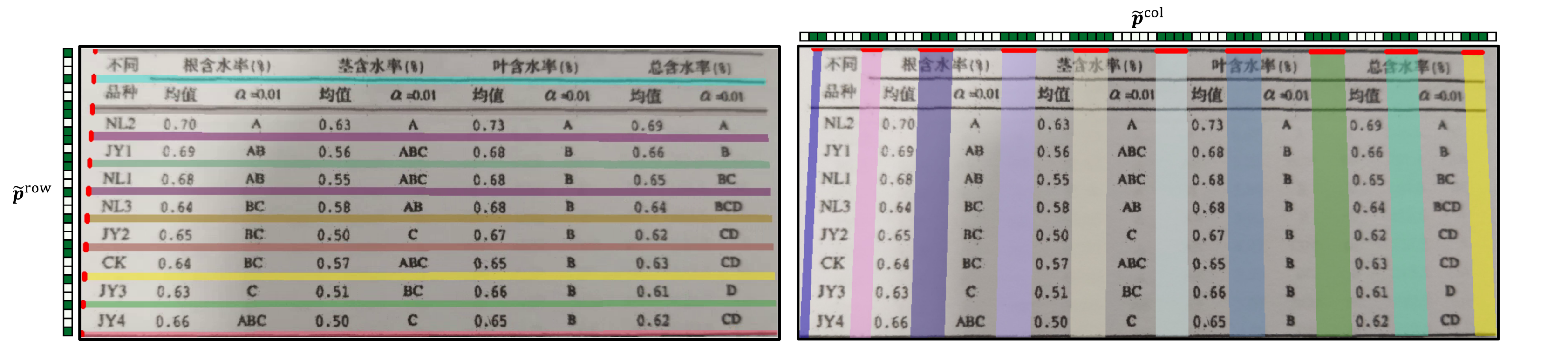}}
		\caption{The visualization of generated table row/column line instance labels $\tilde{\boldsymbol{p}}^{\text{row}}$/$\tilde{\boldsymbol{p}}^{\text{col}}$.}
		\label{start-point}
	\end{figure}

	The $\tilde{\boldsymbol{p}}^{\text{row}}$/$\tilde{\boldsymbol{p}}^{\text{col}}$ is utilized to distinguish different table row/column separation lines. 
	As depicted in Figure~\ref{start-point}, we define $\tilde{\boldsymbol{p}}^{\text{row}}$/$\tilde{\boldsymbol{p}}^{\text{col}}$ as the projection of the start points of table row/column separation lines on the y-axis/x-axis.
	
	\textbf{Label of Merger}
	Since we obtain the label of the splitter, we can partition the table into a series of basic table grids. 
	According to the row/column information provided by the original table structure annotation, we can parse which table grids belong to the same table cell.
	
	\subsection{Implementation Details}
	The ResNet-34~\cite{ResNet} as our backbone is pre-trained on ImageNet~\cite{ImageNet}. 
	The number of feature channels $C$ and $D$ is set to 256 and 512 respectively. 
	The pool size $R\times R$ of RoIAlign in the embedder is set to $3\times 3$. 
	The hyperparameters $\alpha$ and $\gamma$ of sigmoid focal loss $L_{\text{f}}$ are set to 0.25 and 2. The threshold for binarization operations is set to 0.5. 
	
	The training objective of our model is to minimize the table row/column separation line segmentation loss ($\mathscr{L}_{s}^{\text{row}}$/$\mathscr{L}_{s}^{\text{col}}$), the table row/column separation line instance classification loss ($\mathscr{L}_{\text{inst}}^{\text{row}}$/$\mathscr{L}_{\text{inst}}^{\text{col}}$), and the cell merge loss ($\mathscr{L}_{m}$). 
	The objective function for optimization is shown as follows:
	\begin{equation}
		O=\mathscr{L}_{s}^{\text{row}}+\mathscr{L}_{s}^{\text{col}}+\mathscr{L}_{\text{inst}}^{\text{row}}+\mathscr{L}_{\text{inst}}^{\text{col}}+\mathscr{L}_m
	\end{equation}
	
	We employ the ADADELTA algorithm~\cite{ADADELTA} for optimization, with the following hyper parameters: ${\beta _1} = 0.9$, ${\beta _2} = 0.999$ and $\varepsilon  = {10^{ - 9}}$. 
	We set the learning rate using the cosine annealing schedule~\cite{sgdr} as follows:
	\begin{equation}
		{\eta}_t = {\eta _{min}} + \frac{1}{2}({\eta _{max}} - {\eta _{min}})(1 + \cos (\frac{{{T_{cur}}}}{{{T_{max }}}}\pi ))
	\end{equation}
	where ${\eta}_t$ is the updated learning rate. 
	${\eta _{min}}$ and ${\eta _{max}}$ are the minimum learning rate and the initial learning rate, respectively. 
	$T_{cur}$ and $T_{max}$ are the current number of iterations and the maximum number of iterations, respectively. 
	Here we set ${\eta _{min}} = 10^{ - 6}$ and ${\eta _{max}} = 10^{ - 4}$.
	
	{In our experiments, we do not train the model using the training sets of all the datasets, but rather train it on the training sets of each dataset and test it on the test sets of each dataset.}
	Following the SEM~\cite{SEMv1}, SEMv2 is trained with table images in original size on the SciTSR and PubTabNet. 
	However, in the case of the iFLYTAB dataset, we resize the table images using a randomly selected ratio within the range of $[0.8, 1.2]$. 
	Different from the training strategy employed by TSRFormer~\cite{TSRFormerv1}, which follows a multi-stage process, we train the SEMv2 in an end-to-end manner.
	{
		We initialize SEMv2 with the model trained on the iFLYTAB dataset and then fine-tune it on the cTDaR TrackB1-Historical dataset.
		We crop table regions from original images in the WTW dataset for both training and testing.}
	Our training setup includes a single NVIDIA TESLA V100 GPU with 32GB RAM memory and a batch size of 8 for the SciTSR, PubTabNet and cTDaR.
	For iFLYTAB and WTW, we utilize 8 NVIDIA TESLA V100 GPUs with 32GB RAM memory and a batch size of 48.
	The whole framework was implemented using PyTorch.
	
	\subsection{Visualization}
	\begin{figure*}[t]
		\centerline{\includegraphics[width=1\linewidth]{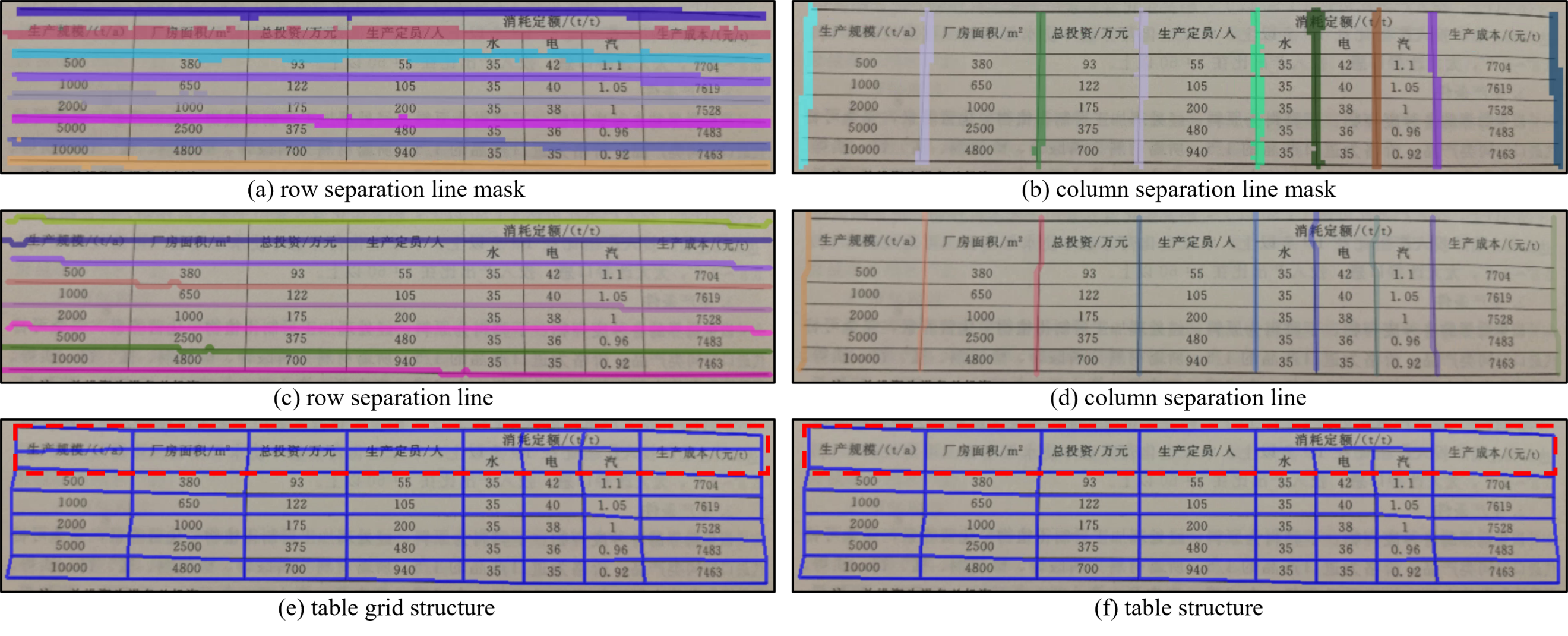}}
		\caption{The visualization of predicted results by SEMv2. The red dash boxes in (e-f) denote the difference between the table grid structure and table structure.}
		\label{visualization_model}
	\end{figure*}
	
	In this section, we visualize the predicted results of both the splitter and the merger to show how SEMv2 recovers the table structure.
	In our work, we formulate table separation line detection as an instance segmentation task, as exemplified in Figure~\ref{visualization_model}(a-b), where different colors represent SEMv2's predictions for distinct instances of table separation lines.
	By finding the maximum score on each column/row of the table row/column separation line mask, we can obtain the table row/column separation lines, as depicted in Figure~\ref{visualization_model}(c-d).
	As illustrated in Figure~\ref{visualization_model}(e), the table grid structure can be derived by intersecting the table row/column separation lines. 
	Through the merger, the spanning cells structure can be restored, ultimately obtaining the final table structure, as shown in the red dashed boxes in Figure~\ref{visualization_model}(f).
	
	\subsection{Ablation Study}
	
	\begin{table}[t]
		\centering
		\caption{Comparison among systems from T1 to T6. ``InstSeg'' means the Instance Segmentation method which is the proposed splitter in this paper. ``SemaSeg'' means the Semantics Segmentation method which follows the splitter in SEM~\cite{SEMv1}. ``ParaDec'' means the Parallel Decoder which is the proposed merger in this paper. ``SeqDec'' means the Sequence Decoder which follows the merger in SEM~\cite{SEMv1}.}
		\label{ablation_system}
		{
			\begin{tabular}{cccccc}
				\toprule
				\multirow{2}{*}{System} & \multicolumn{2}{c}{Splitter} & \multirow{2}{*}{Gather} & \multicolumn{2}{c}{Merger} \\ \cmidrule(lr){2-3} \cmidrule(lr){5-6}
				& InstSeg               & SemaSeg               &                         & ParaDec               & SeqDec                \\ \midrule
				T1                      & \cmark & \xmark & \cmark   & \cmark & \xmark \\
				T2                      & \xmark & \cmark & \cmark   & \cmark & \xmark \\
				T3                      & \cmark & \xmark & \xmark   & \cmark & \xmark \\
				T4                      & \cmark & \xmark & \cmark   & \xmark & \xmark \\
				T5                      & \cmark & \xmark & \cmark   & \xmark & \cmark \\ 
				T6						& \xmark & \cmark & \xmark   & \xmark & \cmark \\ \bottomrule
			\end{tabular}
		}
	\end{table}
	
	\begin{table}[t]
		\centering
		\caption{{The performance by using different splitter modules on the SciTSR and iFLYTAB datasets. The system T6 actually is our baseline system SEM without the textual branch. \textbf{Bold} indicates the best result.}}
		\label{ablation_study_splitter}
		{
			\begin{tabular}{cccccccc}
				\toprule
				\multirow{2}{*}{System} & \multicolumn{3}{c}{SciTSR} & \multicolumn{4}{c}{iFLYTAB}      \\ \cmidrule(lr){2-4} \cmidrule(lr){5-8}
				& P     & R     & F1   & P    & R    & F1   & TEDS-Struct \\ \midrule
				T1      & 99.3  & 99.2  & \textbf{99.3} & 93.8 & 93.3 & \textbf{93.5} & \textbf{92.0}  \\
				T2      & 99.1  & 98.5  & 98.8  & 81.9 & 72.7 & 77.0 & 75.8  \\
				T6		& 99.0  & 98.0  & 98.5  & 81.7	& 74.5 & 78.0 & 75.9  \\	\bottomrule
			\end{tabular}
		}
	\end{table}
	
	\begin{figure}[t]
		\centerline{\includegraphics[width=1.\linewidth]{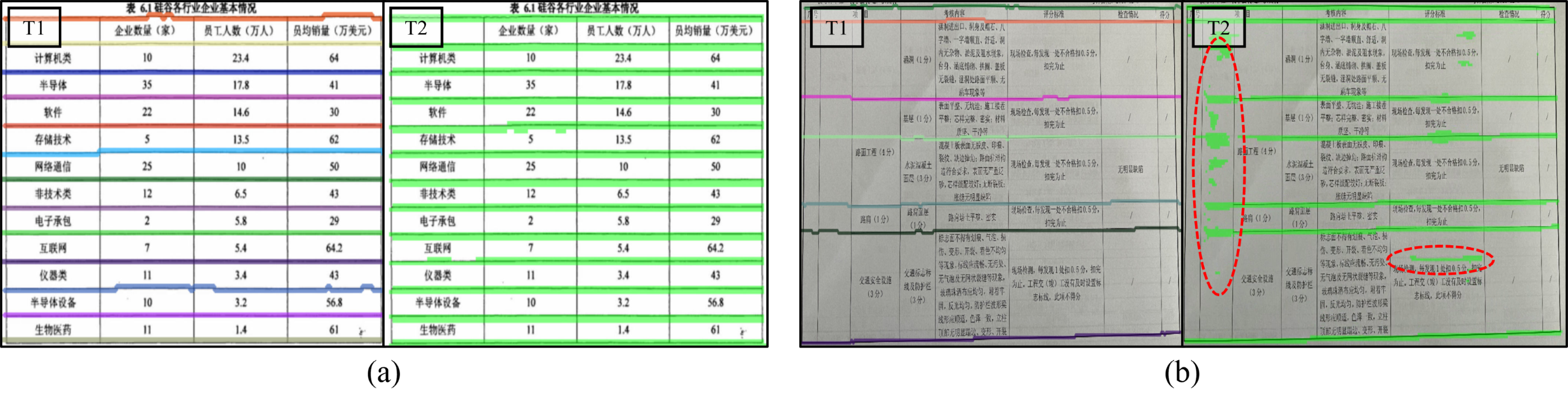}}
		\caption{The segmentation results from the splitters of the designed systems T1 and T2. (a) and (b) correspond to images extracted from digital documents and camera-captured documents, respectively.}
		\label{split_comparsion}
	\end{figure}
	
	\begin{table}[t]
		\centering
		\caption{{The performance on the employment of the gather module on the SciTSR and iFLYTAB datasets. The system T6 actually is our baseline system SEM without the textual branch. \textbf{Bold} indicates the best result.}}
		\label{ablation_study_gather}
		{
			\begin{tabular}{ccccccccc}
				\toprule
				\multirow{2}{*}{System} & \multicolumn{3}{c}{SciTSR} & \multicolumn{4}{c}{iFLYTAB} \\ \cmidrule(lr){2-4} \cmidrule(lr){5-8}
				& P     & R     & F1    & P    & R    & F1   & TEDS-Struct  \\ \midrule
				T1      & 99.3  & 99.2  & \textbf{99.3} & 93.8 & 93.3 & \textbf{93.5} & \textbf{92.0}  \\
				T3      & 98.6  & 98.2  & 98.4 & 91.1 & 89.8 & 90.4 & 89.3  \\
				T6		& 99.0  & 98.0  & 98.5 & 81.7 & 74.5 & 78.0 & 75.9  \\ \bottomrule
			\end{tabular}
		}
	\end{table}
	
	\begin{table}[t]
		\centering
		\caption{{The performance and Frames Per Second (FPS) by using different merger modules on the SciTSR dataset. The system T6 actually is our baseline system SEM without the textual branch. \textbf{Bold} indicates the best result and \underline{underline} indicates the second best.}}
		\label{ablation_study_merger}
		{
			\begin{tabular}{ccccc}
				\toprule
				\multirow{2}{*}{System} & \multicolumn{4}{c}{SciTSR} \\ 
				\cmidrule(lr){2-5}
				& P     & R     & F1   & FPS \\ \midrule
				T1      & 99.3  & 99.2  & \textbf{99.3} & \underline{7.3} \\
				T4      & 99.1  & 97.4  & 98.2 & \textbf{8.9} \\
				T5      & 99.3  & 99.2  & \underline{99.2} & 2.9 \\
				T6	    & 99.0  & 98.0  & 98.5 & 3.1 \\	\bottomrule
			\end{tabular}
		}
	\end{table}
	
	To verify the effectiveness of each component, we conduct ablation experiments through several designed systems as shown in Table~\ref{ablation_system}. 
	The model is not modifed except for	the component being tested. 
	As depicted in Table~\ref{ablation_system}, T6 represents our baseline system, essentially representing SEM~\cite{SEMv1} without the textual branch. 
	On the other hand, T1 corresponds to our proposed SEMv2. 
	As shown in Table~\ref{ablation_study_splitter}, by comparing T1 and T6, it is evident that our approach exhibits superiority over SEM in terms of both performance and efficiency.
	
	\textbf{The effectiveness of the splitter}. 
	In contrast to the majority of previous ``split-and-merge'' based methods, we formulate table separation line detection in the ``split'' stage as an instance segmentation task rather than a semantic segmentation. 
	To evaluate the efficacy of our proposed splitter, we devis the systems T1 and T2 as shown in Table~\ref{ablation_system}. 
	Specifically, T1 represents our proposed SEMv2, whereas T2 replaces the splitter with the one used in the previous state-of-the-art method, SEM\cite{SEMv1}. 
	As shown in Table~\ref{ablation_study_splitter}, although the performance of T1 is only marginally better than T2 on datasets comprising axis-aligned scanned PDF documents (e.g., SciTSR), T1 exhibits a significantly superior performance on the iFLYTAB dataset. 
	This is because the iFLYTAB dataset features camera-captured images with severe deformation, bending, or occlusions. 
	Additionally, we present the segmentation results from splitters of both T1 and T2 in Figure~\ref{split_comparsion}. 
	It can be seen that both T1 and T2 can get high-quality masks on the digital document, but on the camera-captured document, the prediction result of T2 is relatively lower. 
	It’s very hard for the mask-to-line post-processing module to handle such low-quality masks well. 
	In contrast, our instance segmentation based method can easily obtain the shape of the table separation line in a row-wise or column-wise manner, which is more robust to such challenging tables.
	
	
	\textbf{The effectiveness of the Gather}. 
	To illustrate the effectiveness of the Gather module, as shown in Table~\ref{ablation_system}, we designed a system T3, which eliminates RowGather/ColGather, and obtains $\boldsymbol{G}^\text{row}$/$\boldsymbol{G}^\text{col}$ by calculating the column/row mean value of $\boldsymbol{F}$. 
	As shown in Table~\ref{ablation_study_gather}, T1 outperforms T3 by a large margin on both SciTSR and iFLYTAB datasets, which demonstrates the effectiveness of the Gather module for capturing horizontal/vertical visual clues.
	
	\textbf{The efficiency of the merger}. 
	As shown in Table~\ref{ablation_system}, we design the systems T1, T4 and T5 that employ different mergers. 
	T4 eliminates the merger, while T5 substitutes it with the one utilized in SEM\cite{SEMv1}. 
	As shown in Table~\ref{ablation_study_merger}, though T4 exhibits slightly higher Frames Per Second (FPS) than T1, its performance deteriorates significantly as it disregards table cells that span multiple rows or columns. 
	The comparison between T1 and T4 also illustrates the indispensability of the merger. 
	The merger in SEM predicts the merging of grids in a step-by-step manner. 
	As the number of table cells increases, the costed time in the decoding stages of T5 rise, causing the FPS of T5 to be much lower than T1 and T4.
	
	\subsection{Comparison with State-of-the-art Methods}
\begin{figure*}[t]
		\centerline{\includegraphics[width=1.\linewidth]{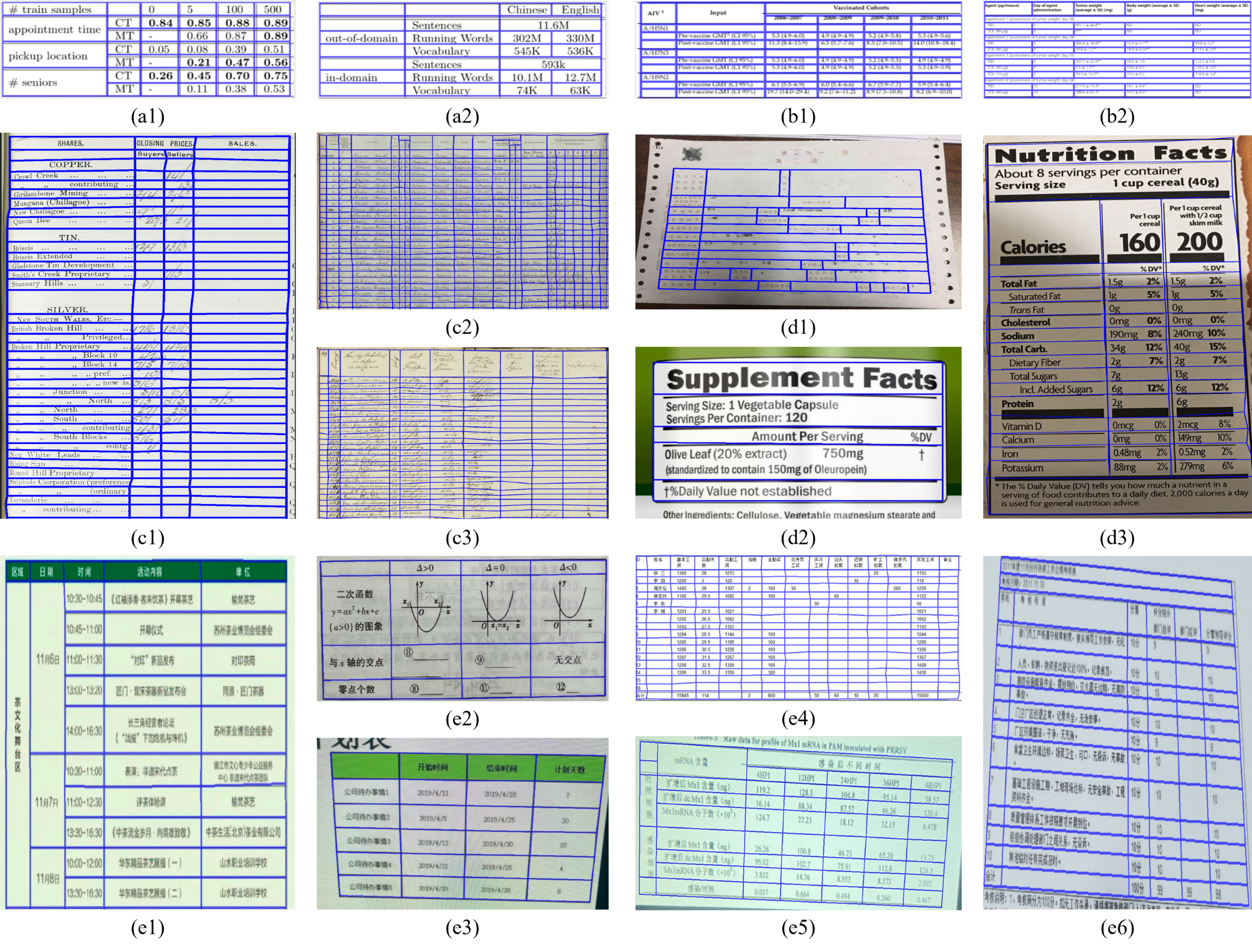}}
		\caption{{Some structure recognition results of SEMv2. a* refer to SciTSR, b* refer to PubTabNet, c* refer to cTDaR, d* refer to WTW, e* refer to iFLYTAB. (e1) refers to the digital wired table, (e2)$\sim$(e3) refer to camera-captured wired tables, (e4) refers to the digital wireless table, (e5)$\sim$(e6) refer to camera-captured wireless tables.}}
		\label{visualization}
	\end{figure*}

	{We compare our method with other state-of-the-art methods on five TSR datasets, including SciTSR, PubTabNet, cTDaR, WTW and iFLYTAB. The results are shown in Tables~\ref{comparsion_sota}~\ref{comparsion_cTDaR}~\ref{comparsion_WTW}~\ref{comparsion_iFLYTAB_SEM}. Furthermore, we visualize the prediction results of our method as shown in Figure~\ref{visualization}. {Finally, we discuss the differences between SEMv2 and other methods that adhere to the ``split-and-merge'' principle.}}
	
	\begin{table*}[t]
		\centering
		\caption{{Comparsion with state-of-the-art methods. \textbf{Bold} indicates the SOTA and \underline{underline} indicates the second best.}}
		\label{comparsion_sota}
		{
			\begin{tabular}{lccccccccc}
				\toprule
				\multirow{2}{*}{Method} & \multicolumn{3}{c}{SciTSR} & \multicolumn{3}{c}{SciTSR-COMP} & \multicolumn{2}{c}{PubTabNet} \\ \cmidrule(lr){2-4} \cmidrule(lr){5-7} \cmidrule(lr){8-9} 
				& P    & R    & F1   & P    & R    & F1   & TEDS & TEDS-Struct   \\ \midrule
				EDD~\cite{EDD}          		 & -    & -    & -       		  & -    & -    & -    & 88.3      & -    \\
				TabStructNet~\cite{TabStructNet} & 92.7 & 91.3 & 92.0             & 90.9 & 88.2 & 89.5 & -    	   & 90.1 \\
				GraphTSR~\cite{SciTSR}     		 & 95.9 & 94.8 & 95.3 			  & 96.4 & 94.5 & 95.5 & -    	   & -    \\
				SEM~\cite{SEMv1}          	     & 97.7 & 96.5 & 97.1 			  & 96.8 & 94.7 & 95.7 & 93.7 	   & 96.3 \\
				LGPMA~\cite{LGPMA}        		 & 98.2 & 99.3 & 98.8 			  & 97.3 & 98.7 & 98.0 & 94.6 	   & 96.7 \\
				RobusTabNet~\cite{RobustTabNet} & 99.4 & 99.1 & \underline{99.3} & 99.0 & 98.4 & \underline{98.7} & -    & \underline{97.0} \\
				TSRFormer~\cite{TSRFormerv1}       & 99.5 & 99.4 & \textbf{99.4}    & 99.1 & 98.7 & \textbf{98.9}    & -    & \textbf{97.5}   \\ \midrule
				SEMv2        					 & 99.3 & 99.2 & \underline{99.3} & 98.7 & 98.6 & \underline{98.7} & -    & \textbf{97.5}   \\ \bottomrule
			\end{tabular}
		}
	\end{table*}
	
	\begin{table}[t]
		\centering
		\caption{{The performance comparsion on the cTDaR TrackB1-Historical dataset. \textbf{Bold} indicates the SOTA}}
		\label{comparsion_cTDaR}
		\begin{tabular}{lccccc}
			\toprule
			\multirow{2}{*}{Team} & \multicolumn{2}{c}{IoU=0.6} & \multicolumn{2}{c}{IoU=0.7} & \multirow{2}{*}{WAvg.F1} \\ \cmidrule(lr){2-3} \cmidrule(lr){4-5}
			& P    & R    & P    & R    &      \\ \midrule
			HCL IDORAN & 25.0 & 5.0  & 23.0 & 4.0  & 6.0  \\
			NLPR-PAL   & 76.0 & 83.0 & 69.0 & 75.0 & 48.0 \\
			SEMv2      & 89.2 & 82.8 & 88.3 & 79.2 & \textbf{67.5} \\ \bottomrule
		\end{tabular}
	\end{table}
	
	\begin{table}[t]
		\centering
		\caption{{The performance comparsion on the WTW dataset. \textbf{Bold} indicates the SOTA.}}
		\label{comparsion_WTW}
		\begin{tabular}{lccc}
			\toprule
			Method & P    & R    & F1   \\ \midrule
			Cycle-CenterNet~\cite{WTW} & 93.3 & 91.5 & 92.4 \\
			TSRFormer~\cite{TSRFormerv1} & 93.7 & 93.2 & 93.4 \\ 
			SEMv2					   & 93.8 & 93.4 & \textbf{93.6} \\	\bottomrule
		\end{tabular}
	\end{table}

	\begin{table}[t]
		\centering
		\caption{{The F1 score comparsion on different categories of the WTW dataset. \textbf{Bold} indicates the SOTA.}}
		\label{comparsion_WTW_subset}
		\begin{tabular}{lccccccc}
			\toprule
			\multirow{2}{*}{Method} & \multirow{2}{*}{Simple} & \multirow{2}{*}{Inclined} & \multirow{2}{*}{Curved} & Occluded & Extreme & \multirow{2}{*}{Overlaid} & Multi \\
			&      &      &      & and blurred & aspect ratio &      & color and grid   \\ \midrule
			Cycle-CenterNet & \textbf{99.3}   & \textbf{97.7}     & 76.1   & 77.4    & 91.9           & \textbf{84.1}  & \textbf{93.7} \\
			SEMv2			& 97.9   & 96.0     & \textbf{87.1}   & \textbf{85.7}    & \textbf{96.1}  & 75.1     & 92.1 \\	\bottomrule
		\end{tabular}
	\end{table}
	
	\begin{table}[t]
		\centering
		\caption{{The performance comparsion with SEM on the iFLYTAB dataset. \textbf{Bold} indicates the SOTA.}}
		\label{comparsion_iFLYTAB_SEM}
		\begin{tabular}{lcccccc}
			\toprule
			Method & P    & R    & F1   & TEDS-Struct & $\text{GriTS}_{\text{Con}}$ & $\text{GriTS}_{\text{Top}}$    \\ \midrule
			SEM    & 81.7 & 74.5 & 78.0 & 75.9 & 79.8 & 82.3 \\
			SEMv2  & 93.8 & 93.3 & \textbf{93.5} & \textbf{92.0} & \textbf{94.2} & \textbf{94.6} \\ \bottomrule
		\end{tabular}
	\end{table}
	
	\begin{figure}[t]
		\centerline{\includegraphics[width=.45\linewidth]{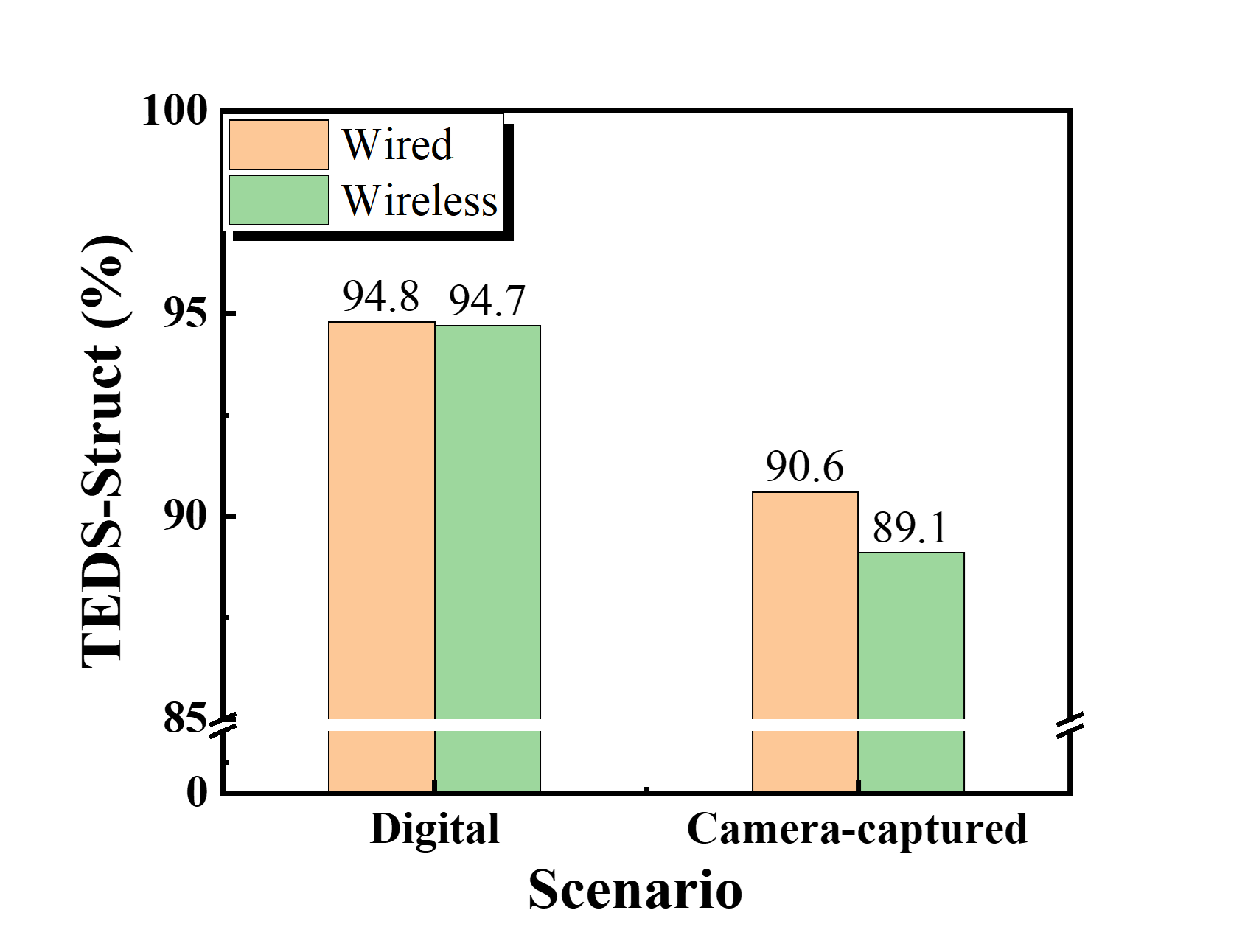}}
		\caption{The performance of SEMv2 on a series of iFLYTAB subsets.}
		\label{comparsion_subiFLYTAB}
	\end{figure}
	
	\textbf{SciTSR and PubTabNet}
	As shown in Table~\ref{comparsion_sota}, our method achieves competitive performance compared to state-of-the-art methods.
	The test dataset of SciTSR exhibits a few annotation errors. Following the RobusTabNet~\cite{RobustTabNet}, we manually rectify these annotation errors. However, it may result in an unfair comparison with other methods.
	The more equitable comparison can be observed on the PubTabNet.
	It is worth noting that the LGPMA~\cite{LGPMA} emerged as the winner of the ICDAR 2021 Competition on Scientific Literature Parsing, Task B. 
	
	{\textbf{cTDaR TrackB1-Historical} To verify the eﬀectiveness of our approach on tabular objects in various scenes, we conduct experiments on the cTDaR TrackB1-Historical dataset. The table images in this dataset are historical documents with handwritten.
		As shown in Table~\ref{comparsion_cTDaR}, we compare the SEMv2 with the participant teams of ICDAR 2019 Competition on Table Detection and Recognition, TrackB2~\cite{Icdar19}. Our method outperforms other teams by a large margin.}
	
	{\textbf{WTW} To verify the eﬀectiveness of our approach on wired distorted/curved tabular objects in wild scenes, we also conduct experiments on the WTW dataset.
		The results in Table~\ref{comparsion_WTW} show that our method achieves comparable performance with state-of-the-art methods. }
	{To investigate the performance of SEMv2 in table structure recognition across various scenarios, we conducted tests on a series of subsets divided by WTW as shown in Table~\ref{comparsion_WTW_subset}. For polygon detection methods like Cycle-CenterNet~\cite{WTW}, SEMv2 performs better in scenarios such as curved, occluded and blurred, and extreme aspect ratio. However, in overlaid scenes, we found that SEMv2's performance was inferior. In the overlaid subset, most of the table images exhibit significant angular rotation, which reduces the overall performance of table structure recognition. How to enhance the performance of SEMv2 in scenarios with obvious angle rotation is also a direction for our future research.}
	
	\textbf{iFLYTAB}
	The iFLYTAB dataset is distinct from SciTSR and PubTabNet in that it encompasses table images taken by cameras. 
	These images are typically accompanied by intricate backgrounds and non-rigid deformations, which makes them more challenging. 
	It is worth noting that detection-based methods (e.g. TabStructNet~\cite{TabStructNet}, LGPMA~\cite{LGPMA}) are subject to the constraints that tables are free of visual rotation or perspective transformation. 
	This condition is difficult to satisfy when table images are camera-captured. 
	Therefore, as shown in Table~\ref{comparsion_iFLYTAB_SEM}, we reimplement the closest method, SEM~\cite{SEMv1}, for a fair comparison on the iFLYTAB dataset. 
	Since the iFLYTAB does not provide the text content annotation, we remove the textual feature in our reimplemented SEM.
	It can be seen that SEMv2 outperforms SEM by a large margin. 
	As depicted in Figure~\ref{comparsion_subiFLYTAB}, we also evaluate the performance of SEMv2 on different subsets of the iFLYTAB dataset. 
	Notably, the model's performance is comparatively lower in the camera-captured scenarios due to the suboptimal image qualities.
	Furthermore, the model performance is less satisfactory on wireless tables than on wired ones, as the former lacks crucial visual information.
	
	{\textbf{Split-and-merge} Previous works such as TSRFormer~\cite{TSRFormerv1}, SEM~\cite{SEMv1}, SPLERGE~\cite{SPLERGE} and RobusTabNet~\cite{RobustTabNet} all follow the ``split-and-merge'' principle. The main difference between these methods is found in their individual split stages. SEM, SPLERGE and RobusTabNet design the splitter based on the semantic segmentation, which requires a complex mask-to-line algorithm to extract table row/column separation lines from the predicted masks. In contrast, SEMv2 designs an instance segmentation-based method in the split stage, which predicts a mask for each table row/column separation line. Considering that most table row/column separation lines are spread through horizontal/vertical direction, we propose a simple “mask-to-line” algorithm, as shown in Figure~\ref{post-processing}, that can accurately extract table row/column separation lines. TSRFormer formulates separation line prediction as a line regression problem instead of an image segmentation problem. Specifically, TSRFormer predicts reference points and using a DETR decoder to regress line coordinates. As shown in Table~\ref{comparsion_WTW}, TSRFormer and SEMv2 are comparable in performance.}
	
	\section{Error Analysis}
\begin{figure*}[t]
		\centerline{\includegraphics[width=1\linewidth]{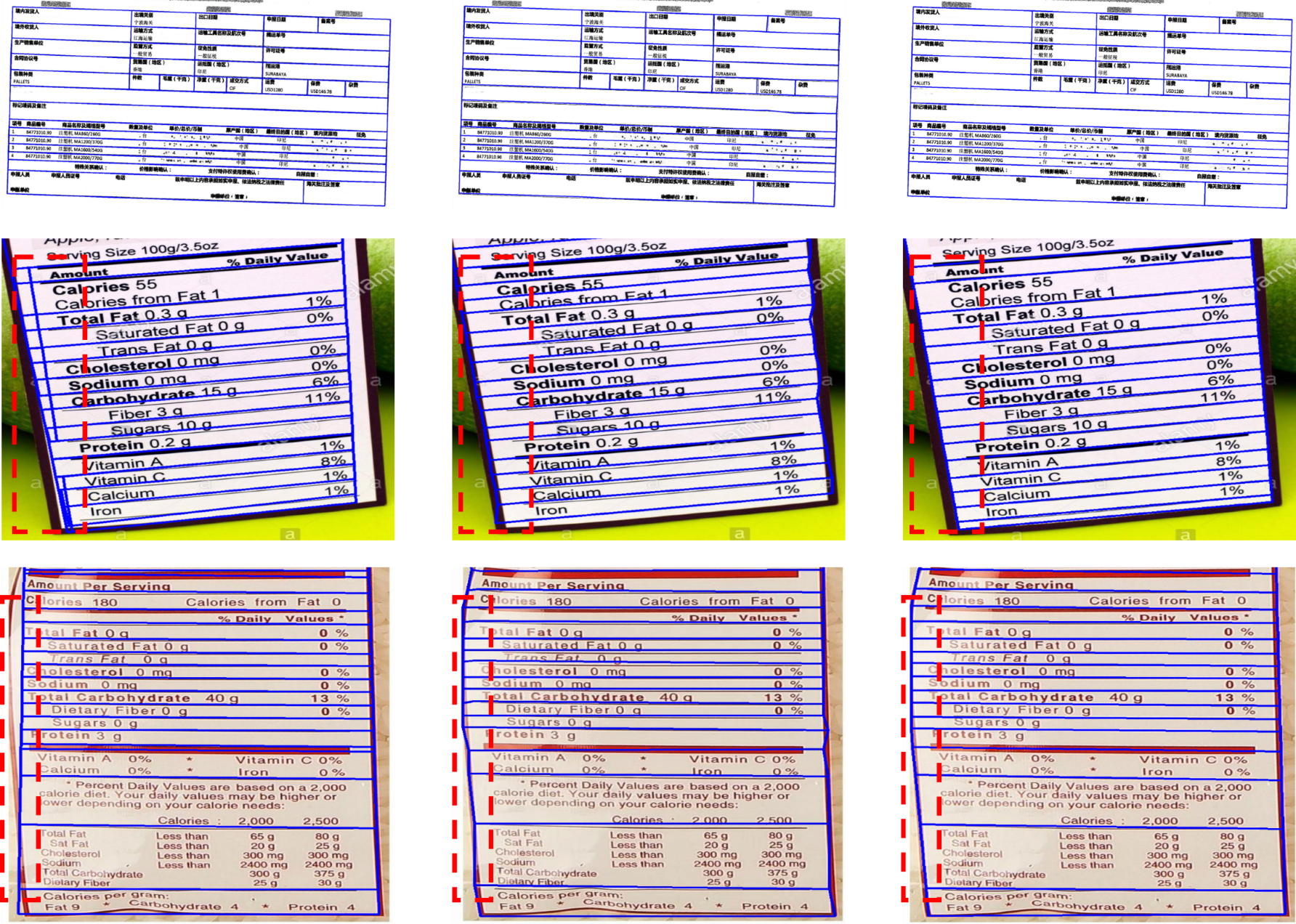}}
		\caption{{Some table structure recognition results obtained by TSRFormer and SEMv2 on the WTW dataset.
				\textbf{First Column} the table structure predictions from the TSRFormer.
				\textbf{Second Column} the table structure predictions from the SEMv2.
				\textbf{Third Column} the ground truth table structure.
				The red dash boxes denote the different prediction regions between TSRFormer and SEMv2.}}
		\label{semv2_vs_tsrformer}
	\end{figure*}

	{\textbf{WTW} In this section, we present the table structure recognition results obtained by TSRFormer~\cite{TSRFormerv1} and SEMv2 on the iFLYTAB dataset, as shown in Figure~\ref{semv2_vs_tsrformer}. 
	TSRFormer follows the ``split-and-merge'' approachs~\cite{SEMv1,RobustTabNet} and proposes a line regression method to detect table separation lines in the split stage. 
	As shown in Table~\ref{comparsion_WTW}, TSRFormer performs similarly to SEMv2 in terms of performance. 
	As shown in the first row of Figure~\ref{semv2_vs_tsrformer}, both methods can handle scanned documents well. 
	In the second and third rows of Figure~\ref{semv2_vs_tsrformer}, SEMv2 performs slightly better in some natural scenes. 
	For example, in the third row, SEMv2 can achieve more accurate results for some curved table separation lines.}
	
\begin{figure*}[t]
		\centerline{\includegraphics[width=1\linewidth]{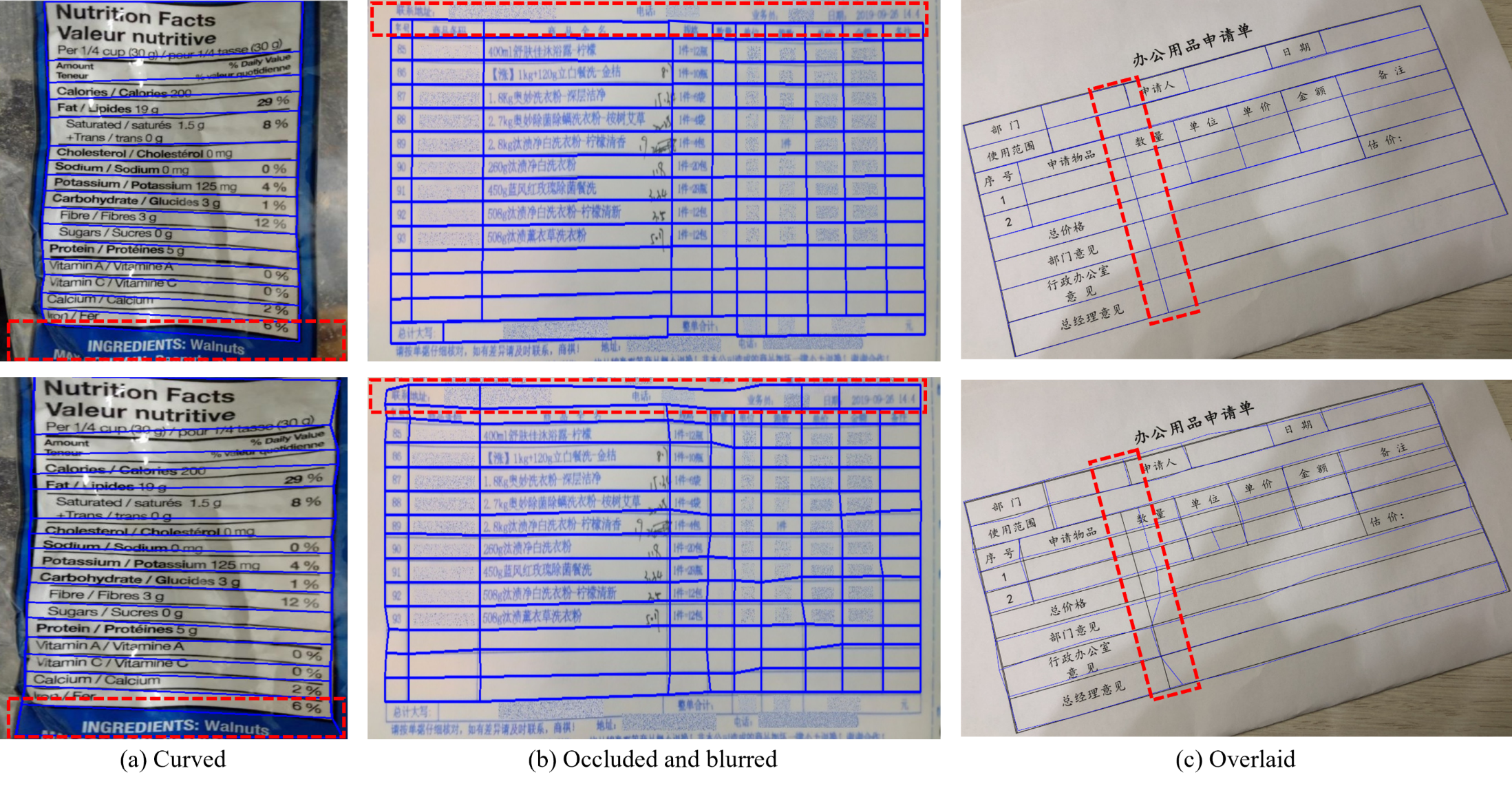}}
		\caption{{Some table structure recognition results obtained by SEMv2 on different categories of the WTW dataset. 
				\textbf{First Row} the ground truth table structure. 
				\textbf{Second Row} the table structure predictions from the SEMv2.
				The red dash boxes denote the different prediction regions between ground truth and SEMv2.}}
		\label{wtw_error_sample}
	\end{figure*}

	{Additionally, as shown in Figure~\ref{wtw_error_sample}, we also present three scenarios where SEMv2 performs poorly on the WTW test set. As demonstrated in Figure~\ref{wtw_error_sample}(a-b), for text outside the table area, SEMv2 mistakenly judges it as a row of the table, resulting in lower accuracy. For tables with obvious angle rotation, although SEMv2 has a certain capability to handle it, there is still a gap compared to polygon detection methods~\cite{WTW}. How to improve the model's performance in scenarios with significant angle rotation will be a direction for our future research.}
	
	\begin{figure*}[t]
		\centerline{\includegraphics[width=1\linewidth]{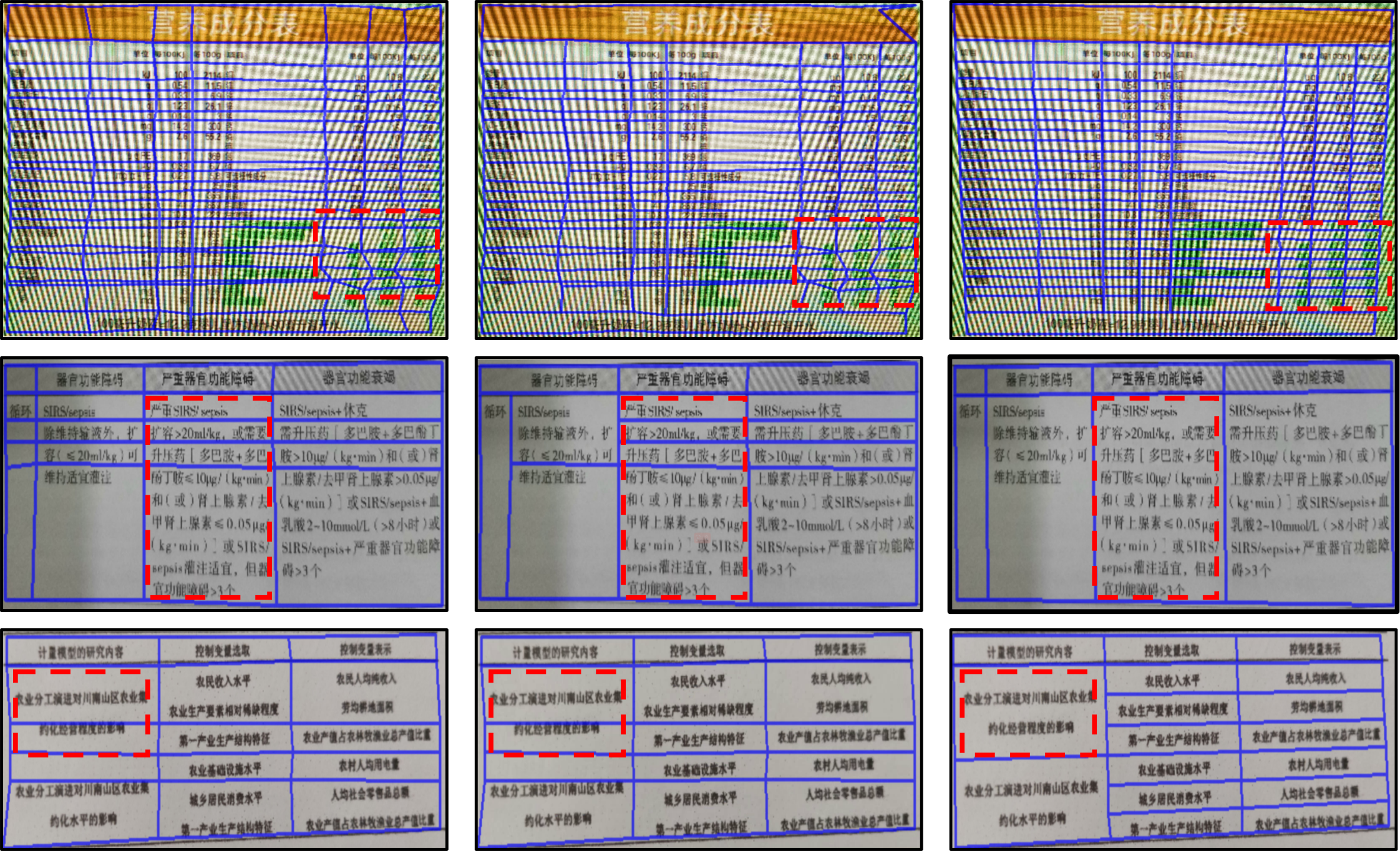}}
		\caption{{Some incorrect table structure recognition results by SEMv2 on the iFLYTAB dataset.
			\textbf{First Column} the predictions of the table grid structure from the splitter.
			\textbf{Second Column} the final predictions of the table structure from the merger.
			\textbf{Third Column} the ground truth table structure.
			The red dash boxes denote the incorrect prediction regions.}}
		\label{error_analysis}
	\end{figure*}
	
	\textbf{iFLYTAB}
	In this section, we present erroneous table structure recognition results obtained by SEMv2 on the iFLYTAB dataset, as depicted in Figure~\ref{error_analysis}. 
	As illustrated in the first row of Figure~\ref{error_analysis}, the iFLYTAB dataset comprises screen-captured table images, characterized by pronounced moiré patterns, which adversely affect the performance of our approach.
	The second and third rows of Figure~\ref{error_analysis} showcase table cells within the iFLYTAB dataset that harbor multi-line content.
	As explicated by SEM~\cite{SEMv1}, the model necessitates a comprehensive comprehension of the textual elements within the table images to facilitate more precise predictions.
	To enhance the efficiency of the model, SEMv2 refrains from introducing the textual branch, in contrast to its predecessor SEM, which consequently results in underperformance when confronted with table cells containing multi-line content.
	Devising efficient strategies to enable the model to understand the textual content in the table images and predict accurate table structure will be our future work.
	
	\section{Conclusion and Future Work}
	In this paper, we propose a novel method for tackling the problem of table structure recognition, SEMv2. It mainly contains three components including splitter, embedder and merger. Distinct from previous methods in the ``split'' stage, SEMv2 aims to distinguish each table line and formulate the table line detection as an instance segmentation task. The ablation experiment also demonstrates that our proposed splitter is more robust to tables in various scenarios. Moreover, we propose a parallel decoder based on conditional convolution for the merger, which significantly boosting the model's efficiency. To comprehensively evaluate the SEMv2, we also present a more challenging dataset for table structure recognition, named iFLYTAB. We collect and annotate 17291 tables (both wired and wireless type tables) in various scenarios such as camera-captured photos, scanned documents, etc. Some table images in iFLYTAB have intricate backgrounds and non-rigid deformations. The comprehensive experiments on the publicly available datasets (e.g. SciTSR, PubTabNet) and the proposed iFLYTAB dataset illustrate that the SEMv2 achieves a state-of-the-art performance for table structure recognition. We hope our proposed iFLYTAB dataset can further advance future research on table structure recognition.
	
	{SEMv2 achieves state-of-the-art performance on table structure recognition, but it still has some limitations. 
	{As discussed in the error analysis section, the model's recognition capabilities are notably compromised when dealing with images that not only exhibit significant rotational distortion but also suffer from poor image quality.} 
	Additionally, SEMv2 performs poorly in predicting the structure of cells with multi-line text due to the lack of textual information.
	To make SEMv2 more robust, we will study an efficient multi-modal table structure recognition scheme in our future work. This scheme will fully utilize the textual, visual, and layout information in table images to complete the prediction of table structure recognition. Furthermore, we will also study the application of computer vision techniques in table structure recognition to eliminate moiré patterns, severe image distortion, and other challenges, thereby improving the model's prediction accuracy.}

	{\small \bibliography{reference}}
	
\end{document}